\documentclass[10pt,twocolumn,letterpaper]{article}

\usepackage{iccv}              

\usepackage{comment}
\usepackage{amsmath}
\usepackage{amssymb}
\usepackage{graphicx}
\usepackage{caption}
\usepackage{bm}
\usepackage{breqn}
\usepackage{array}
\usepackage{siunitx}
\usepackage{times}
\usepackage{algorithm}
\usepackage{algpseudocode}
\usepackage{bbm}
\usepackage{multirow}
\usepackage{arydshln}
\usepackage{makecell}

\definecolor{iccvblue}{rgb}{0.21,0.49,0.74}
\usepackage[pagebackref,breaklinks,colorlinks,allcolors=iccvblue]{hyperref}
\def\paperID{****} 
\def\confName{ICCV}
\def\confYear{2025}

\title{CE-FAM: Concept-Based Explanation via Fusion of Activation Maps}

\author{Michihiro Kuroki $\quad$ Toshihiko Yamasaki\\
The University of Tokyo\\
{\tt\small kuroki@cvm.t.u-tokyo.ac.jp}
}

\begin{document}
\maketitle
\begin{abstract}
Although saliency maps can highlight important regions to explain the reasoning behind image classification in artificial intelligence (AI), the meaning of these regions is left to the user's interpretation. In contrast, concept-based explanations decompose AI predictions into human-understandable concepts, clarifying their contributions. However, few methods can simultaneously reveal what concepts an image classifier learns, which regions are associated with them, and how they contribute to predictions. \\
\indent We propose a novel concept-based explanation method, Concept-based Explanation via Fusion of Activation Maps (CE-FAM). It employs a branched network that shares activation maps with an image classifier and learns to mimic the embeddings of a Vision and Language Model (VLM). The branch network predicts concepts in an image, and their corresponding regions are represented by a weighted sum of activation maps, with weights given by the gradients of the concept prediction scores. Their contributions are quantified based on their impact on the image classification score. Our method provides a general framework for identifying the concept regions and their contributions while leveraging VLM knowledge to handle arbitrary concepts without requiring an annotated dataset. Furthermore, we introduce a novel evaluation metric to assess the accuracy of the concept regions. Our qualitative and quantitative evaluations demonstrate our method outperforms existing approaches and excels in zero-shot inference for unseen concepts.
\end{abstract}    
\section{Introduction}
\label{sec:intro}
\captionsetup[subfigure]{aboveskip=1pt}
\captionsetup[subfigure]{justification=centering}
\setlength\textfloatsep{3pt}
Artificial Intelligence (AI) has been rapidly advancing and is increasingly being integrated into various aspects of daily life, with the potential to further enhance our everyday experiences. While AI is expected to play a crucial role in complex domains such as autonomous driving and healthcare, its black-box nature poses the risk of unexpected behaviors, which may result in critical failures. To ensure safe and effective human-AI collaboration, enhancing AI transparency and interpretability is essential.

\begin{figure}[t]
\centering
\includegraphics[width=0.80\linewidth]{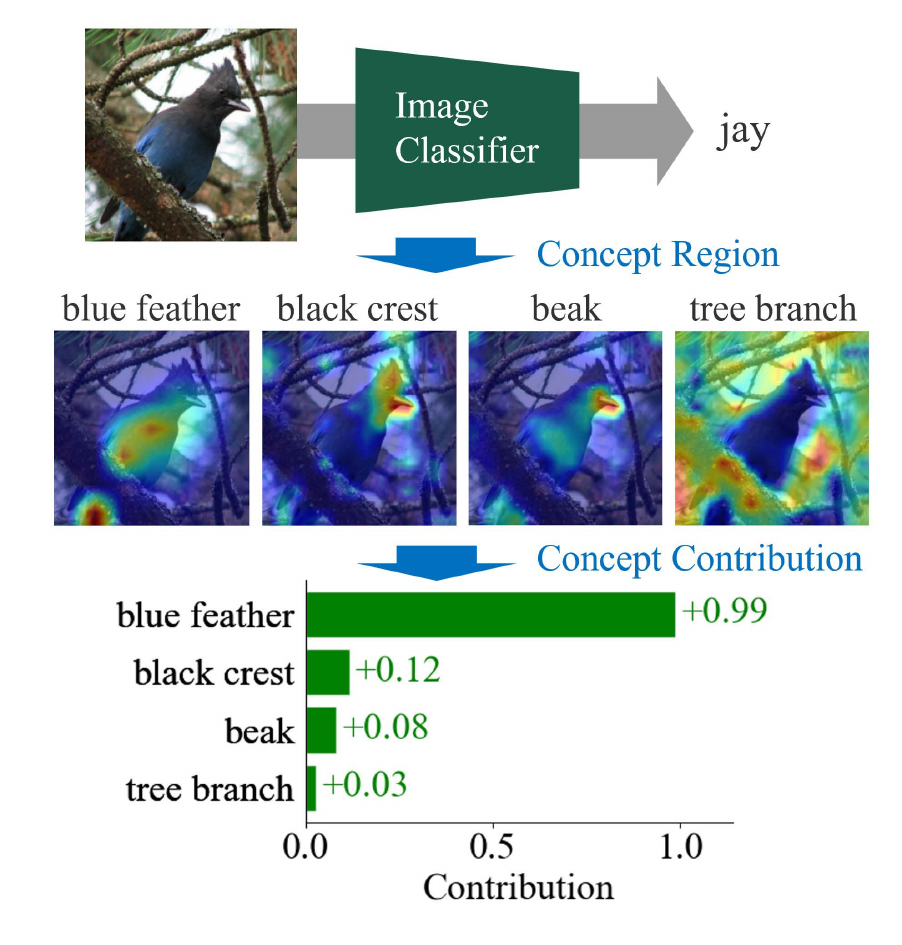}
\vspace{-5pt}
\caption{Overview of the proposed method, showing the concepts underlying classification predictions, their corresponding regions, and their contributions to the final prediction.}
\label{fig:1}
\end{figure}

\begin{figure*}[t]
\centering
	\begin{tabular}{p{1\linewidth}}
		\centering
    		\begin{minipage}[t]{0.11\linewidth}
      		\subcaption*{}
   		\end{minipage}
		\hspace{15pt}
    		\begin{minipage}[t]{0.22\linewidth}
      		\centering
      		\subcaption*{CE-FAM (Ours)}
    		\end{minipage}
		\hspace{15pt}
    		\begin{minipage}[t]{0.22\linewidth}
      		\centering
      		\subcaption*{CLIP-Dissect~\cite{CLIP-dissect}}
    		\end{minipage}
		\hspace{15pt}
    		\begin{minipage}[t]{0.22\linewidth}
      		\centering
      		\subcaption*{WWW~\cite{WWW}}
	    	\end{minipage} \\
		\vspace{-10pt}
    		\begin{minipage}[t]{0.11\linewidth}
      		\centering
      		\subcaption*{Target: hair}
      		\includegraphics[width=0.82\linewidth]{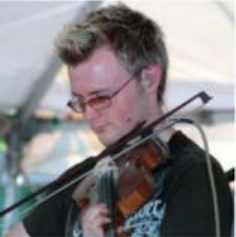}
   		\end{minipage}
		\hspace{15pt}
    		\begin{minipage}[t]{0.11\linewidth}
      		\centering
      		\subcaption*{L4 (0.32)}
      		\includegraphics[width=0.82\linewidth]{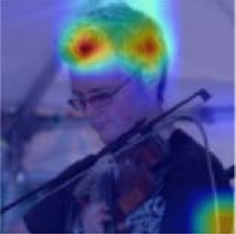}
    		\end{minipage}
    		\begin{minipage}[t]{0.11\linewidth}
      		\centering
      		\subcaption*{L2 (0.27)}
      		\includegraphics[width=0.82\linewidth]{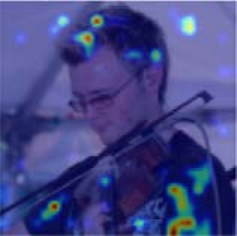}
    		\end{minipage}
		\hspace{15pt}
    		\begin{minipage}[t]{0.11\linewidth}
      		\centering
      		\subcaption*{L4 \#1633 (4.91)}
      		\includegraphics[width=0.82\linewidth]{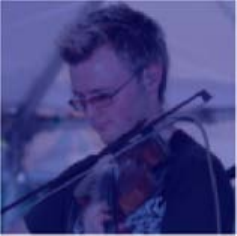}
    		\end{minipage}
    		\begin{minipage}[t]{0.11\linewidth}
      		\centering
      		\subcaption*{L4 \#730 (4.50)}
      		\includegraphics[width=0.82\linewidth]{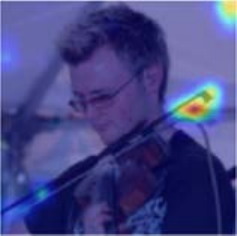}
    		\end{minipage}
		\hspace{15pt}
    		\begin{minipage}[t]{0.11\linewidth}
      		\centering
      		\subcaption*{L4 \#173 (1.16)}
      		\includegraphics[width=0.82\linewidth]{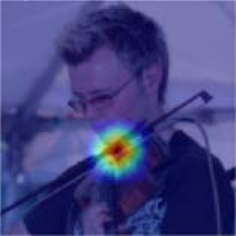}
    		\end{minipage}
    		\begin{minipage}[t]{0.11\linewidth}
      		\centering
      		\subcaption*{L4 \#1633 (0.66)}
      		\includegraphics[width=0.82\linewidth]{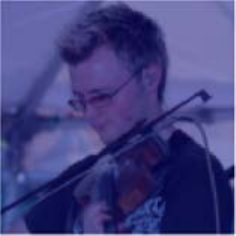}
	    	\end{minipage} \\
		\vspace{-8pt}
    		\begin{minipage}[t]{0.11\linewidth}
      		\centering
      		\subcaption*{Target: hair}
      		\includegraphics[width=0.82\linewidth]{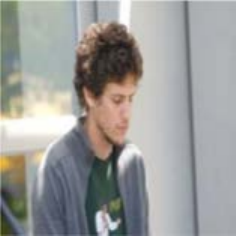}
   		\end{minipage}
		\hspace{15pt}
    		\begin{minipage}[t]{0.11\linewidth}
      		\centering
      		\subcaption*{L4 (0.55)}
      		\includegraphics[width=0.82\linewidth]{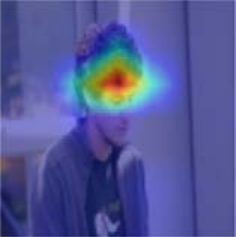}
    		\end{minipage}
    		\begin{minipage}[t]{0.11\linewidth}
      		\centering
      		\subcaption*{L3 (0.33)}
      		\includegraphics[width=0.82\linewidth]{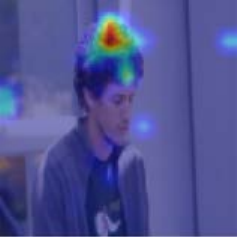}
    		\end{minipage}
		\hspace{15pt}
    		\begin{minipage}[t]{0.11\linewidth}
      		\centering
      		\subcaption*{L4 \#1633 (4.91)}
      		\includegraphics[width=0.82\linewidth]{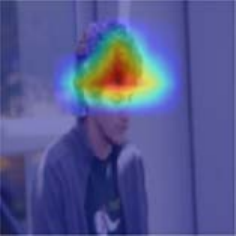}
    		\end{minipage}
    		\begin{minipage}[t]{0.11\linewidth}
      		\centering
      		\subcaption*{L4 \#730 (4.50)}
      		\includegraphics[width=0.82\linewidth]{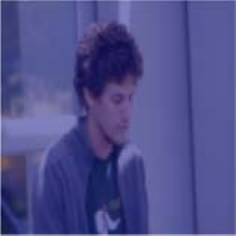}
    		\end{minipage}
		\hspace{15pt}
    		\begin{minipage}[t]{0.11\linewidth}
      		\centering
      		\subcaption*{L4 \#173 (1.16)}
      		\includegraphics[width=0.82\linewidth]{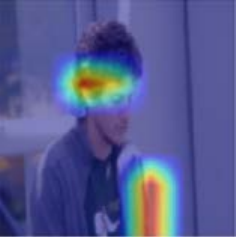}
    		\end{minipage}
    		\begin{minipage}[t]{0.11\linewidth}
      		\centering
      		\subcaption*{L4 \#1633 (0.66)}
      		\includegraphics[width=0.82\linewidth]{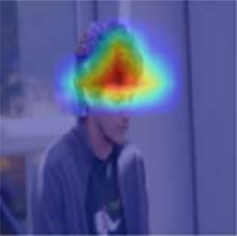}
	    	\end{minipage} \\
		\vspace{-8pt}
    		\begin{minipage}[t]{0.11\linewidth}
      		\centering
      		\subcaption*{Target: green}
      		\includegraphics[width=0.82\linewidth]{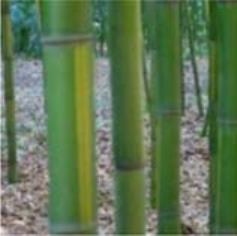}
   		\end{minipage}
		\hspace{15pt}
    		\begin{minipage}[t]{0.11\linewidth}
      		\centering
      		\subcaption*{L3 (0.76)}
      		\includegraphics[width=0.82\linewidth]{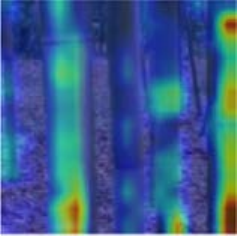}
    		\end{minipage}
    		\begin{minipage}[t]{0.11\linewidth}
      		\centering
      		\subcaption*{L4 (0.61)}
      		\includegraphics[width=0.82\linewidth]{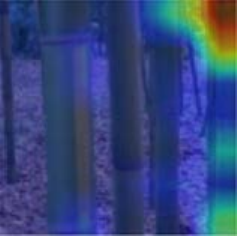}
    		\end{minipage}
		\hspace{15pt}
    		\begin{minipage}[t]{0.11\linewidth}
      		\centering
      		\subcaption*{L3 \#134 (3.87)}
      		\includegraphics[width=0.82\linewidth]{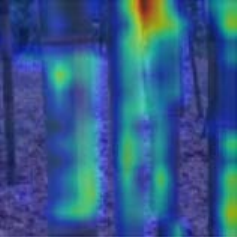}
    		\end{minipage}
    		\begin{minipage}[t]{0.11\linewidth}
      		\centering
      		\subcaption*{L3 \#968 (3.83)}
      		\includegraphics[width=0.82\linewidth]{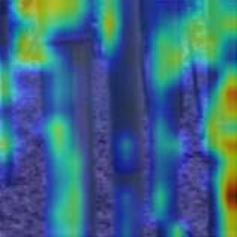}
    		\end{minipage}
		\hspace{15pt}
    		\begin{minipage}[t]{0.11\linewidth}
      		\centering
      		\subcaption*{L4 \#1715 (1.92)}
      		\includegraphics[width=0.82\linewidth]{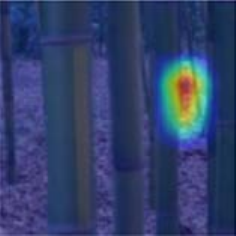}
    		\end{minipage}
    		\begin{minipage}[t]{0.11\linewidth}
      		\centering
      		\subcaption*{L4 \#2046 (1.86)}
      		\includegraphics[width=0.82\linewidth]{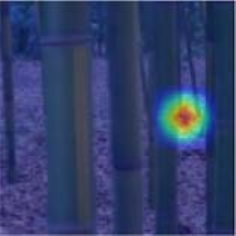}
	    	\end{minipage} 
    	\end{tabular}
\vspace{-8pt}
\caption{Comparison of regions corresponding to target concepts. Existing methods~\cite{CLIP-dissect, WWW} present the top 2 candidate neurons (specified by layer name and index) associated with the concept, ranked by similarity score in descending order, and visualize their activation maps. Due to space limitations, "Layer" is abbreviated as "L," and the similarity score (hereafter, association score) is indicated in parentheses.}
\label{fig:2}
\end{figure*}
Explainable AI (XAI) has gained significant attention for enhancing AI transparency. For example, saliency maps highlight important regions in image classification by visualizing pixel-wise contributions. Grad-CAM~\cite{GradCAM} is a widely used method that generates a saliency map by weighting and summing activation maps in the final layer of a Convolutional Neural Network (CNN). It uses the gradients obtained by backpropagation of the class prediction score as the weights. However, the limited expressiveness of saliency maps has recently been recognized as a challenge. Since the interpretation of highlighted regions is left to the user, identifying which specific aspects, such as patterns or shapes, are truly critical remains difficult. In contrast, concept-based explanations have been proposed as a method for identifying the factors that contribute to a prediction. Concepts refer to human-interpretable elements, such as colors, textures, and object parts. For example, Concept Bottleneck Model (CBM)~\cite{CBM} introduces feature representations corresponding to each concept just before the final fully connected layer, and makes predictions based on their linear combination. TCAV~\cite{TCAV} utilizes the Concept Activation Vector (CAV), a vector representation of each concept, to analyze which concepts influence predictions by comparing CAVs with image embeddings from a classifier. Although acquiring feature representations corresponding to each concept has traditionally required fine-grained annotated datasets, handling arbitrary concepts without annotation labels has recently become more feasible by leveraging embeddings from a pre-trained Vision and Language Model (VLM). However, while these methods clarify the contribution of individual concepts by utilizing concept-specific feature representations, they struggle to identify the corresponding regions. Demonstrating where concepts exist and how they contribute to predictions, as shown in Fig.~\ref{fig:1}, is important for human-understandable explanations. However, few methods simultaneously satisfy both requirements. \\
\indent Dissection-based methods associate individual channels of activation maps in CNN layers (referred to as neurons in these methods) with concepts, enabling the identification of corresponding regions. CLIP-Dissect~\cite{CLIP-dissect} utilizes a VLM, such as CLIP~\cite{CLIP}, to compute the similarity score distribution for each concept across all images in a dataset. It then associates each concept with the neuron whose activation intensity best aligns with its similarity score distribution. In WWW~\cite{WWW}, images in which the target neuron is highly activated are sampled, and concepts with an average similarity score computed by CLIP that exceeds an adaptive threshold are linked to that neuron. However, these methods face the issue that the association between concepts and activation maps is a many-to-many relationship. Fig.~\ref{fig:2} visualizes how each concept is associated with each neuron. The target image classifier is ResNet50~\cite{ResNet}, and each neuron is identified by the model's layer name and channel index. Since multiple candidate pairs exist, Fig.~\ref{fig:2} presents the top two results with the highest similarity scores. However, these results indicate that a high similarity score does not necessarily imply the most appropriate pair, and the optimal pair varies depending on the sample. Based on this observation, we suggest that representing a concept using a single activation map has inherent limitations. A similar insight has been reported in previous studies. Net2Vec~\cite{Net2Vec} proposes representing concepts by combining multiple filters. In line with this approach, we investigate the effectiveness of representing concepts through the fusion of activation maps.\\
\indent In this paper, we propose a novel concept-based explanation method, Concept-based Explanation via Fusion of Activation Maps (CE-FAM), which identifies weights for the activation maps to represent each concept, thereby providing concept regions and their contributions. Our method trains a network branching off from the shared activation maps of an image classifier to mimic a VLM's image embedding. The learned embedding enables the prediction of concepts within an image by computing its similarity to the text embeddings of each concept. By backpropagating the concept prediction score, we obtain gradients that are used to weight activation maps for representing each concept. Our approach directly indicates concept regions, and quantifies their contributions by measuring their impact on the classification score. To the best of our knowledge, our method is the first to establish a one-to-one correspondence between concept labels, their associated regions, and their contributions. Additionally, since previous studies have not evaluated the accuracy of the regions corresponding to concepts, this paper introduces a novel evaluation metric and demonstrates the superiority of our method over prior approaches. The contributions of this paper are as follows:
\begin{itemize}
\item We propose an XAI method that provides concept regions and their contributions, offering a more comprehensive and human-interpretable approach. This framework can be generally applied to image classifiers with architectures that utilize activation maps.
\item We propose an evaluation metric for assessing concept regions and conduct evaluations using various combinations of image classifiers, VLMs, and datasets. The results demonstrate that our method outperforms existing methods under various conditions.
\item Experimental results demonstrate that our method achieves superior performance even in a zero-shot setting with respect to concept labels, as it can handle arbitrary concepts without relying on annotated datasets by leveraging VLM knowledge.
\end{itemize}

\section{Related Works}
\label{sec:2_relatework}
\subsection{Saliency-based Explanation}
Saliency maps are a representative XAI method in image classification, visualizing each pixel's contribution to class predictions. They can be categorized into several types. As a method that uses backpropagated class prediction scores~\cite{LRP, IG, SmoothGrad, GuidedBackprop}, Integrated Gradients~\cite{IG} assigns pixel-wise contributions by integrating gradients obtained while transforming an input image into a baseline image. As a method that leverages weighted activation maps in the final CNN layer~\cite{GradCAM, ScoreCAM, CAM, GradCAM++}, Grad-CAM~\cite{GradCAM} weights activation maps using the gradients of the prediction score. As a method that utilizes output changes from perturbed inputs~\cite{RISE, LIME, ExternalPerturbation}, RISE~\cite{RISE} computes a weighted sum of masks, using prediction scores from masked images as weights. Although some methods~\cite{SHAP, SHAPCAM, fastSHAP, VitSHAP} leverage the Shapley value~\cite{ShapleyValue} to ensure fairness in explainability, saliency maps are still limited to highlighting regions, leaving the interpretation of those pixels to the user.

\begin{figure*}[t]
\centering
	\begin{tabular}{p{0.92\linewidth}}
		\centering
    		\begin{minipage}[t]{0.48\linewidth}
      		\centering
      		\includegraphics[width=1\linewidth]{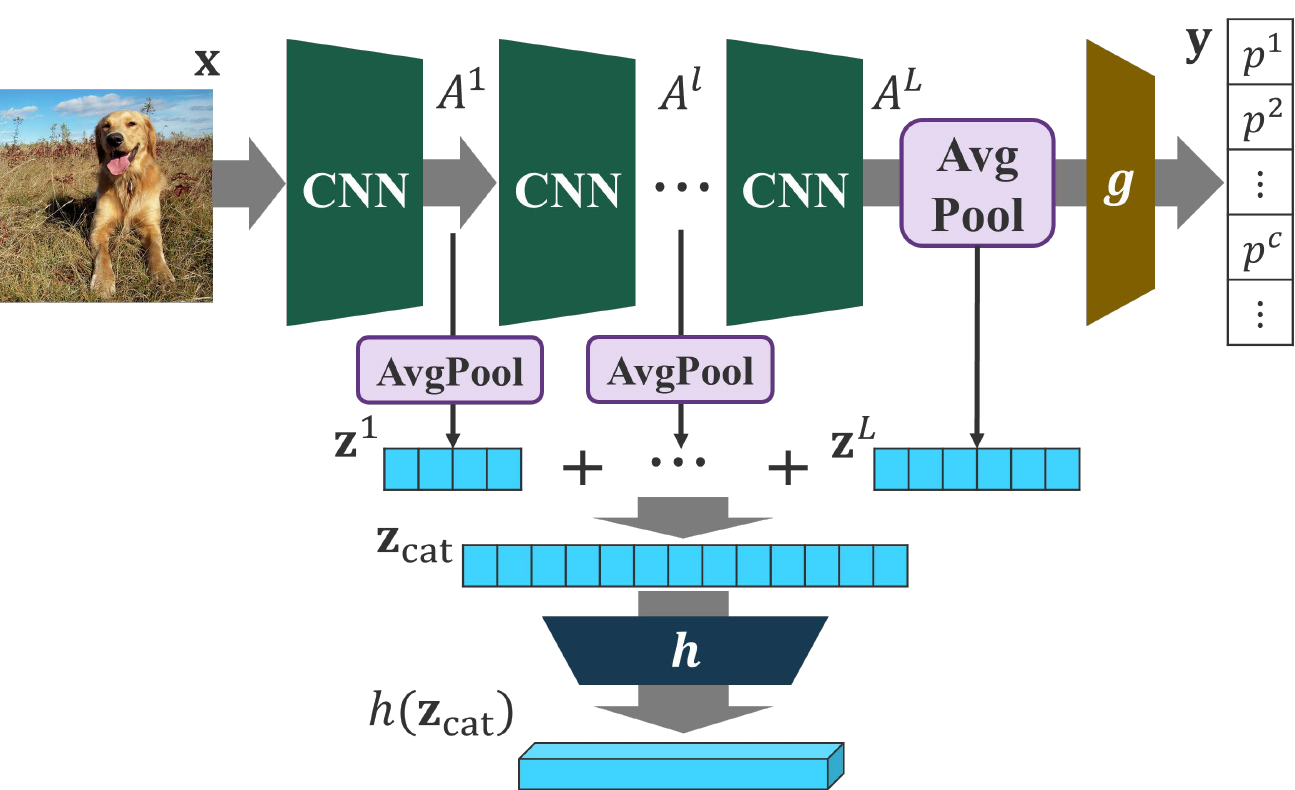}
   		\end{minipage}
		\hspace{12pt}
    		\begin{minipage}[t]{0.45\linewidth}
      		\centering
      		\includegraphics[width=1\linewidth]{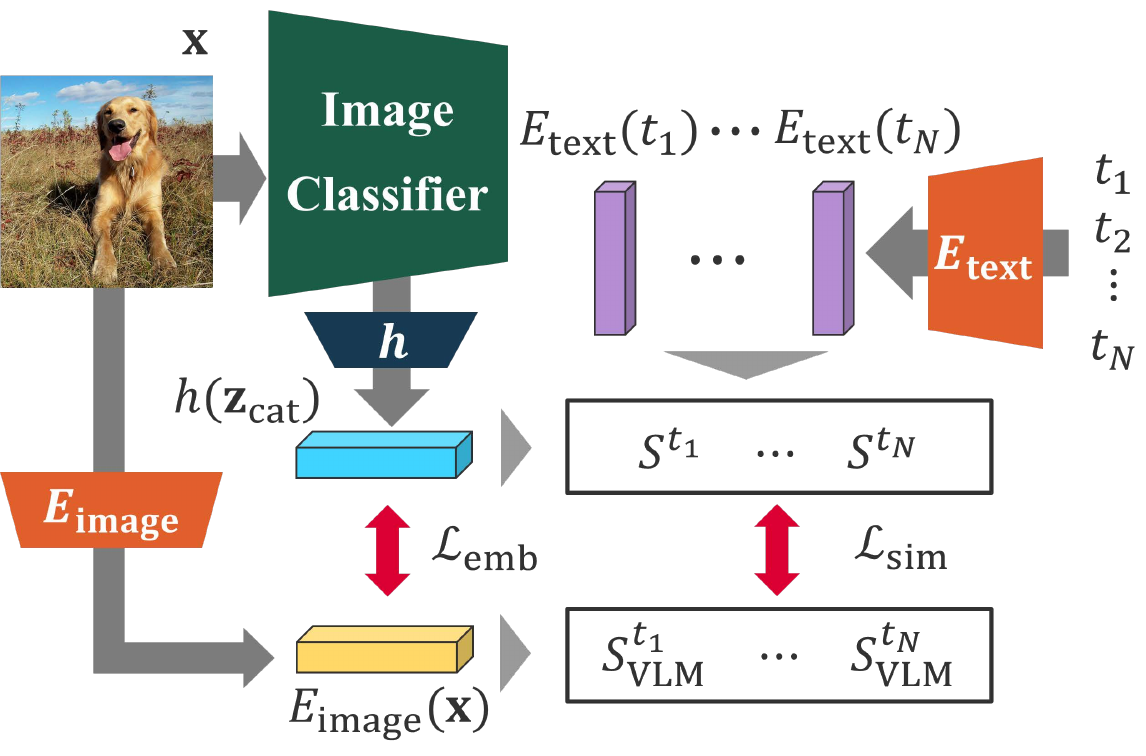}
    		\end{minipage}
    	\end{tabular}
\vspace{-15pt}
\caption{(left) Extraction of the image embedding from the embedding vector $\mathbf{z}_{\textrm {cat}}$ using a translator function $h$. (right) Concept learning framework that leverages the image embedding $E_{\textrm {image}}(\mathbf{x})$ and text embedding $E_{\textrm {text}}(t)$ provided by a VLM.}
\label{fig:3}
\end{figure*}
\subsection{Concept-based Explanation}
\noindent {\bf Conventional approaches.} \indent Concept-based explanations aim to enhance human understanding of the factors contributing to a model's prediction. A concept refers to human-interpretable elements, such as colors, textures, object parts, or entire objects. In Concept Bottleneck Model (CBM)~\cite{CBM}, a layer composed of concept-aligned features is introduced before the final fully connected layer, and predictions are made based on their linear combination. TCAV~\cite{TCAV} employs Concept Activation Vector (CAV), which represents a concept, to analyze which concepts are influential during prediction by measuring their similarity to the gradients of an image classifier's embedding. However, these methods do not identify the regions corresponding to the concepts. Network Dissection~\cite{NetworkDissection} can localize a concept by associating it with a single channel of activation maps using a segmentation mask for each concept. The channel is linked to a concept if the Intersection over Union (IoU) between its highly activated region and the corresponding mask exceeds a threshold. However, some methods~\cite{Net2Vec, LoCE, CLM, glca} propose combining multiple channels with weights because a single channel is insufficient to represent a concept. Net2Vec~\cite{Net2Vec} trains a linear model to align the weighted activation maps with a segmentation mask for each concept, but the number of models grows with the number of concepts. CLM~\cite{CLM} learns concept prediction using an annotated dataset with concept labels and determines weights using gradient-based XAI~\cite{SmoothGrad, VarGrad}. While these methods localize concept regions, they do not clarify how each concept contributes to the final prediction. ECLAD~\cite{ECLAD} clarifies the concept contribution by aggregating sensitivities within its localized regions. CRP~\cite{ConceptRP} quantifies concept contributions by summing relevance scores conditioned on each concept. However, these methods struggle to clarify what each concept represents. As described above, identifying a concept by its label, region, and contribution remains challenging. Moreover, the above methods are limited by their reliance on fine-grained annotated datasets, making annotation costs a significant burden. 
\vskip.5\baselineskip
\noindent {\bf VLM-driven approaches.} \indent Recently, methods leveraging the encoded representations of VLMs~\cite{CLIP, SigLIP, BLIP2} have been proposed to acquire concept representations without requiring manual annotations. For example, Label-free CBM~\cite{Labelfree} trains a function that maps an image classifier's embeddings to concept-aligned feature representations by referencing concept similarity scores computed by a VLM across multiple images. CounTEX~\cite{CounTEX} identifies CAVs based on changes in CLIP's embedding space corresponding to each concept and utilizes them to analyze their impact on predictions. However, these methods struggle to localize specific regions corresponding to concepts. CRAFT~\cite{CRAFT}, which employs vectors obtained through non-negative matrix factorization as CAVs, identifies relevant regions in an image while illustrating representative image samples. However, this approach makes it difficult for humans to interpret concept labels in textual form. Methods~\cite{MILAN, FALCON, CLIP-dissect, WWW} that associate concepts with activation maps in CNN layers can identify concept regions. CLIP-Dissect~\cite{CLIP-dissect} uses the similarity between CLIP's image and text encodings as a concept prediction score. It computes the distribution of concept prediction scores across all images and compares them with the distribution of activation intensities in the classifier’s activation maps. Based on the similarity of these distributions, corresponding concept-neuron pairs are identified. Similarly, WWW~\cite{WWW} samples images in which the target neuron is highly activated and leverages CLIP's concept prediction scores for those images to perform pairing. It also computes each neuron's contribution to the predictions using a Shapley value~\cite{ShapleyValue} approximation to estimate the importance of each concept. Although WWW provides a comprehensive and human-interpretable explanation, the association between concepts and activation maps is sometimes insufficient, as illustrated in Fig.~\ref{fig:2}. In our study, by representing a concept as a weighted fusion of the activation maps, we achieve a one-to-one correspondence between concepts and regions.
\section{Proposed Method}
\label{sec:3_method}
\setlength{\abovedisplayskip}{5pt}
\setlength{\belowdisplayskip}{5pt}
\subsection{Preliminary}
\label{sec:preliminary}
Grad-CAM~\cite{GradCAM} is widely used in XAI for image classification, and our proposed method extends it for concept prediction. Let $\mathbf{x}\in \mathbb{R}^{3\times H\times W}$ denote an input image, and let $\mathbf{z}^L\in \mathbb{R}^d$ denote the image embedding vector extracted from the last CNN layer $L$. The output logits $\mathbf{y}$ from a classifier can be expressed using a function $g$, such as a Multi-Layer Perceptron (MLP), as follows:
\begin{equation}
\label{eq1}
\mathbf{y}=g(\mathbf{z}^L).
\end{equation}
Here, we define $p^c$ as the class prediction score for a target class~$c$, obtained by applying the Softmax function to $\mathbf{y}$. In image classifiers, activation maps are generated through the CNN blocks. When $A^L$ denotes those obtained from the last CNN layer, $\mathbf{z}^L$ is generated by applying global average pooling to $A^L$. In Grad-CAM~\cite{GradCAM}, the saliency map $M^c_{\textrm {Grad-CAM}}$ is represented as a weighted sum of $A^L_k$, which denotes the $k$-th channel in the activation maps $A^L$.
\begin{equation}
\label{eq2}
M^c_{\textrm {Grad-CAM}}={\textrm {ReLU}}\Biggl(\sum_k\alpha^c_k A^L_k\Biggr),
\end{equation}
\begin{equation}
\label{eq3}
\alpha^c_k=\frac{1}{\gamma}\sum_i\sum_j\frac{\partial p^c}{\partial A^L_k(i,j)}.
\end{equation}
Here, $\gamma$ denotes the normalization factor for global average pooling, and ${\textrm {ReLU}}(\cdot)$ represents the ReLU activation function. Following the same process as Grad-CAM~\cite{GradCAM}, our proposed method represents the concept region as a weighted sum of the activation maps. Instead of using the gradients with respect to the class prediction score $p^c$ as the weights, our method employs the gradients with respect to the concept prediction scores as the weights.

\begin{table*}[t]
\footnotesize
\centering
	\begin{tabular}{|>{\centering}m{3.0em}>{\centering}m{5.8em}||>{\centering}m{1.8em}>{\centering}m{1.8em}>{\centering}m{1.8em}>{\centering}m{1.8em}>{\centering}m{1.8em}|>{\centering}m{1.8em}>{\centering}m{1.8em}>{\centering}m{1.8em}>{\centering}m{1.8em}>{\centering}m{1.8em}|>{\centering}m{1.8em}>{\centering}m{1.8em}>{\centering}m{1.8em}>{\centering}m{1.8em}>{\centering\arraybackslash}m{1.8em}|}
		\hline
		& & \multicolumn{5}{c|}{EPG~\cite{ScoreCAM}} & \multicolumn{5}{c|}{NRA} & \multicolumn{5}{c|}{Hit Rate}\\
		\cline{3-17}
		Model & Method & Object & Part & Color & Mate-rial & Avg. & Object & Part & Color & Mate-rial & Avg. & Object & Part & Color & Mate-rial & Avg. \\
		\hline
		\multirow{3}{*}{ResNet50} & CLIP-Dissect & 0.197 & 0.060 & 0.156 & {\bf 0.170} & 0.146 & 0.327 & 0.265 & {\bf 0.393} & 0.350 & 0.334 & 0.215 & 0.113 & {\bf 0.270} & 0.198 & 0.199 \\
		& WWW & 0.179 & {\bf 0.069} & 0.098 & 0.124 & 0.117 & 0.322 & 0.230 & 0.296 & 0.265 & 0.278 & 0.154 & 0.052 & 0.185 & 0.063 & 0.114 \\
		& CE-FAM & {\bf 0.233} & 0.067 & {\bf 0.169} & 0.147 & {\bf 0.154} & {\bf 0.459} & {\bf 0.316} & 0.294 & {\bf 0.375} & {\bf 0.361} & {\bf 0.436} & {\bf 0.185} & 0.114 & {\bf 0.255} & {\bf 0.247} \\
		\hline
		\multirow{3}{*}{ResNet18} & CLIP-Dissect & 0.158 & 0.042 & 0.127 & 0.130 & 0.114 & 0.304 & 0.229 & {\bf 0.394} & 0.343 & 0.318 & 0.101 & 0.038 & {\bf 0.254} & 0.140 & 0.133 \\
		& WWW & 0.125 & 0.038 & 0.156 & 0.080 & 0.100 & 0.287 & 0.211 & 0.354 & 0.339 & 0.298 & 0.041 & 0.014 & 0.206 & 0.025 & 0.071\\
		& CE-FAM & {\bf 0.218} & {\bf 0.061} & {0\bf .181} & {\bf 0.148} & {\bf 0.152} & {\bf 0.431} & {\bf 0.294} & 0.316 & {\bf 0.363} & {\bf 0.351} & {\bf 0.378} & {\bf 0.155} & 0.146 & {\bf 0.243} & {\bf 0.230} \\
		\hline
        \end{tabular}
\vspace{-8pt}
\caption{Evaluation results for concept regions using all images from the Broden~\cite{NetworkDissection} dataset. The highest NRA among the top 4 scoring candidates is evaluated. The average value for each concept type is reported, and "Avg." denotes the mean of these per-type averages.}
\label{tab:1}
\end{table*}

\subsection{Concept Learning}
\label{sec:conceptlearning}
To eliminate dependence on annotated datasets when learning concept representations, many studies~\cite{PosthocCBM, FaithfulCBM, Labo} leverage the knowledge of a VLM. The commonly used CLIP consists of an image encoder $E_{\textrm{image}}$ and a text encoder $E_{\textrm{text}}$, enabling it to compute the similarity between image and text embeddings. Here, let $\mathit{X}$ be the set of images used for concept learning, and $\mathit{T}$ be the set of concept texts. The concept in an image is predicted based on the similarity between an image $\mathbf{x} \in \mathit{X}$ and a concept text $t \in \mathit{T}$.  For the text $t$, we apply a template (e.g., ``a photo of \{\ \}\ '') to the concept label, similar to how it is used in CLIP training. In concept learning, many related studies utilize only the embedding vector $\mathbf{z}^L$ before the fully connected layer of an image classifier. However, higher layers lack information on low-level features, raising concerns that learning concept representations associated with them may be difficult. Therefore, the proposed method utilizes the embedding vectors $\mathbf{z}^l$ obtained at each CNN layer $l\in \{1,2, \ldots ,L\}$, allowing the model to incorporate all features from low to high levels.
\begin{equation}
\label{eq4}
\mathbf{z}^l={\textrm {AvgPool}}\bigl(A^l\bigr),
\end{equation}
\begin{equation}
\label{eq5}
\mathbf{z}_{\textrm {cat}}={\textrm {Concat}}\bigl(\mathbf{z}^1, \mathbf{z}^2, \ldots, \mathbf{z}^{L}\bigr).
\end{equation}
${\textrm {AvgPool}}(\cdot)$ denotes the global average pooling over the spatial dimensions, and ${\textrm {Concat}}(\cdot)$ represents the function that concatenates vectors. $\mathbf{z}_{\textrm {cat}}$ is projected into CLIP's embedding space through a translator function $h$, which is a simple MLP. The overview of this extraction of the image embeddings is described in Fig.~\ref{fig:3}~(left). CLIP's text embedding for a concept $t$ is represented as $E_{\textrm {text}}(t)$. It can be compared with the image embedding $h(\mathbf{z}_{\textrm {cat}})$ within CLIP's embedding space. The concept similarity between $\mathbf{x}$ and $t$ is calculated as follows.
\begin{equation}
\label{eq6}
S^t=h(\mathbf{z}_{\textrm {cat}}) \cdot E_{\textrm {text}}(t),
\end{equation}
\begin{equation}
\label{eq7}
S_{\textrm {VLM}}^t=E_{\textrm {image}}(\mathbf{x}) \cdot E_{\textrm {text}}(t).
\end{equation}
This similarity can be interpreted as a prediction score for the concept, where a higher score indicates a higher likelihood of the concept being present. When the concept text set $T$ is predefined, a similarity loss $\mathcal{L}_{\textrm {sim}}$ based on concept prediction can be incorporated. However, as discussed later in the evaluation results in Section~\ref{sec:discussion}, the model can achieve high performance without this loss function. Since concept texts are not used during training, the setting can be considered zero-shot with respect to concept labels. The total loss function is formulated as follows:
\begin{equation}
\label{eq8}
\mathcal{L}_{\textrm {total}}=\mathcal{L}_{\textrm {emb}}+\lambda\mathcal{L}_{\textrm {sim}},
\end{equation}
\begin{equation}
\label{eq9}
\mathcal{L}_{\textrm {emb}}={\textrm {MSE}} \Bigl( h(\mathbf{z}_{\textrm {cat}}) - E_{\textrm {image}}(\mathbf{x}) \Bigr),
\end{equation}
\begin{equation}
\label{eq10}
\mathcal{L}_{\textrm {sim}}={\textrm {MSE}} \Bigl(S^t-S_{\textrm {VLM}}^t\Bigr).
\end{equation}
$\lambda$ is a parameter for adjusting the scale of the loss functions, and ${\textrm {MSE}}(\cdot)$ represents the Mean Squared Error function. An overview of the framework is shown in Fig.~\ref{fig:3}.

\subsection{Concept Region}
\label{sec:conceptregion}
The regions corresponding to the learned concepts are visualized using Grad-CAM~\cite{GradCAM}. Considering that the CNN captures features at different levels in each layer, it is preferable to generate maps for each layer $l$. We denote $R^l_t$ as a weighted map based on $A^l_k$, indicating the region corresponding to concept $t$ in image $\mathbf{x}$.
\begin{equation}
\label{eq11}
R^l_t={\textrm {ReLU}}\Biggl(\sum_k\beta^t_k A^l_k\Biggr),
\end{equation}
\begin{equation}
\label{eq12}
\beta^t_k=\frac{1}{\gamma}\sum_i\sum_j\frac{\partial S^t}{\partial A^l_k(i,j)}.
\end{equation}
For a single concept $t$, the number of candidate maps is equal to the number of layers. To select the most relevant layer, we calculate a layer-wise association score based on the impact on the concept prediction score $S^t$. The layer with the highest association score can be considered the most relevant to predicting the concept.\\
\indent The impact is measured by the degree of decrease in the score $S^t$ when masking the highly important elements of the embedding vector $\mathbf{z}^l$ that constitute each concept. We define the importance of channel $k$ in layer $l$ as $\beta^t_k\mathbf{z}^l_k$. Since channels with low importance introduce noise in predicting the concept, we consider only the top $K$ channels (e.g., $K=20$). In the masking operation, we set the mask value to $m^l=\frac{1}{|\mathcal{C}|} \sum_{k \in \mathcal{C}} \mathbf{z}^l_k$ and replace $\mathbf{z}^l_k$ with $m^l$. $\mathcal{C}$ denotes the set of indices of all channels. The impact score is the area under the curve (AUC) obtained by masking channels one by one in descending order of $\beta^t_k\mathbf{z}^l_k$ and plotting the decrease in the score $S^t$. We define this impact score as the association score and compute it for each layer $l$.
\subsection{Concept Contribution}
\label{sec:conceptcontribution}
Most existing methods compute concept contributions using information from the layer before the final fully connected layer. While this constraint allows for measuring the direct contribution to the prediction, it fails to account for concepts present in lower layers. In contrast, we propose a method that quantifies contributions from any layer. Similar to the impact score for $S^t$, we use an impact score for class prediction score $p^c$ and define it as concept contribution. The calculation method for the impact score is the same as in Section~\ref{sec:conceptregion}, except that the target is $p^c$ instead of $S^t$. An example plot for computing the AUC is shown in Fig.~\ref{fig:5b}. If the score $p^c$ increases due to masking, the impact score may become negative, indicating a negative contribution.
\section{Evaluation and Results}
\label{sec:4_experiment}
\subsection{Experimental Settings}
\noindent {\bf Models.} \indent We used CLIP (ViT-B/16)~\cite{CLIP} as the VLM, as it is commonly used in related studies. For a fair evaluation, we applied the same model to the baseline methods. We also compared our method using SigLIP~\cite{SigLIP}, which is a higher-performance VLM. For the image classifier, we used ResNet50~\cite{ResNet}, a widely used model, and ResNet18 for comparison. While many related studies focus on CNN-based classifiers, we also evaluated our method on Transformer-based classifiers such as ViT-B/16~\cite{ViT}. In this case, the encoder block output is reshaped and treated as activation maps, a transformation commonly used when applying CAM-based methods~\cite{pytorchgradcam}.
\vskip.5\baselineskip
\noindent {\bf Training Settings.} \indent In concept learning, the maximum number of training epochs is 150, and SGD~\cite{SGD} is used as the optimizer. The initial learning rate is set to 0.1 and increased to 0.2 during the warm-up phase, if applied. We also adopt early stopping, and the learning rate is multiplied by 0.1 if the validation loss does not decrease for 4 consecutive epochs. The loss scaling parameter $\lambda$ is set to 0.001.
\vskip.5\baselineskip
\noindent {\bf Dataset.} \indent We used ImageNet~\cite{ImageNet} as the training image dataset $\mathit{X}$ in all experimental settings. For the evaluation of concept regions, we employed Broden~\cite{NetworkDissection} and ImageNet-S~\cite{ImageNetS} because they provide segmentation masks corresponding to concept labels. Broden consists of 63,305 images collected from several datasets~\cite{OpenSurfaces, ADE, PascalContent, PascalPart}. 1,197 unique concept labels are categorized into types such as color, object, part, material, scene, and texture. The scene and texture types were excluded from our evaluation because only label names were provided, and no segmentation masks were available. ImageNet-S is an extension of ImageNet with additional segmentation mask labels.
\vskip.5\baselineskip
\noindent {\bf Baseline Methods.} \indent To evaluate concept regions, we compare existing methods that associate concepts with neurons. Similar to WWW~\cite{WWW}, some methods~\cite{MILAN, FALCON} assign concepts using images where the target neuron is highly activated and refine their descriptions. However, since our evaluation focuses on a unified set of concept labels and WWW~\cite{WWW} is sufficient to evaluate association methods, these methods are excluded from the evaluation. For a fair comparison, we limited the scope to methods using VLMs and selected CLIP-Dissect~\cite{CLIP-dissect} and WWW~\cite{WWW} as baselines.

\subsection{Qualitative Evaluation of Concept Region} 
\label{sec:42}
We used ResNet50~\cite{ResNet} as the target image classifier and performed concept learning using CLIP~\cite{CLIP}. Fig.~\ref{fig:1} shows an example of concept regions and their contributions for a \textit{jay} image in ImageNet~\cite{ImageNet}. For clarity, the selected concepts are those that actually appear in the image. It is evident that each region is clearly highlighted and \textit{blue feather} plays a dominant role in the prediction of \textit{jay}. Next, we compare the validity of the concept regions with those from existing methods~\cite{CLIP-dissect, WWW}. Since these methods can identify multiple candidates, we present the top two results along with their association scores. In our method, a well-defined region is obtained for the concepts of \textit{hair} and \textit{green}. Since the second candidate does not produce a meaningful region, it can be inferred that a specific layer is predominantly responsible for learning a particular concept.
\begin{figure}[t]
	\centering
	\begin{tabular}{p{0.92\linewidth}}
		\centering
    		\begin{minipage}[t]{0.47\linewidth}
      			\centering
      			\includegraphics[width=1\linewidth]{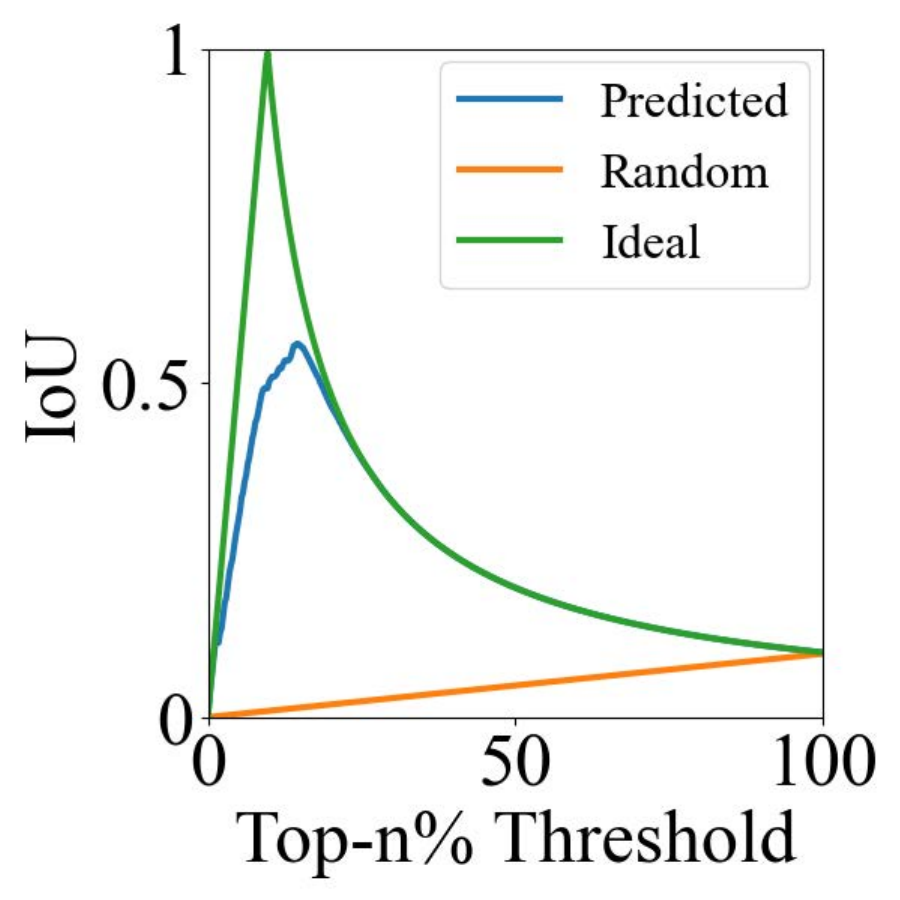}
      			\subcaption{hair (NRA: 0.833)}
			\label{fig:4a}
   		\end{minipage}
    		\begin{minipage}[t]{0.47\linewidth}
      			\centering
      			\includegraphics[width=1\linewidth]{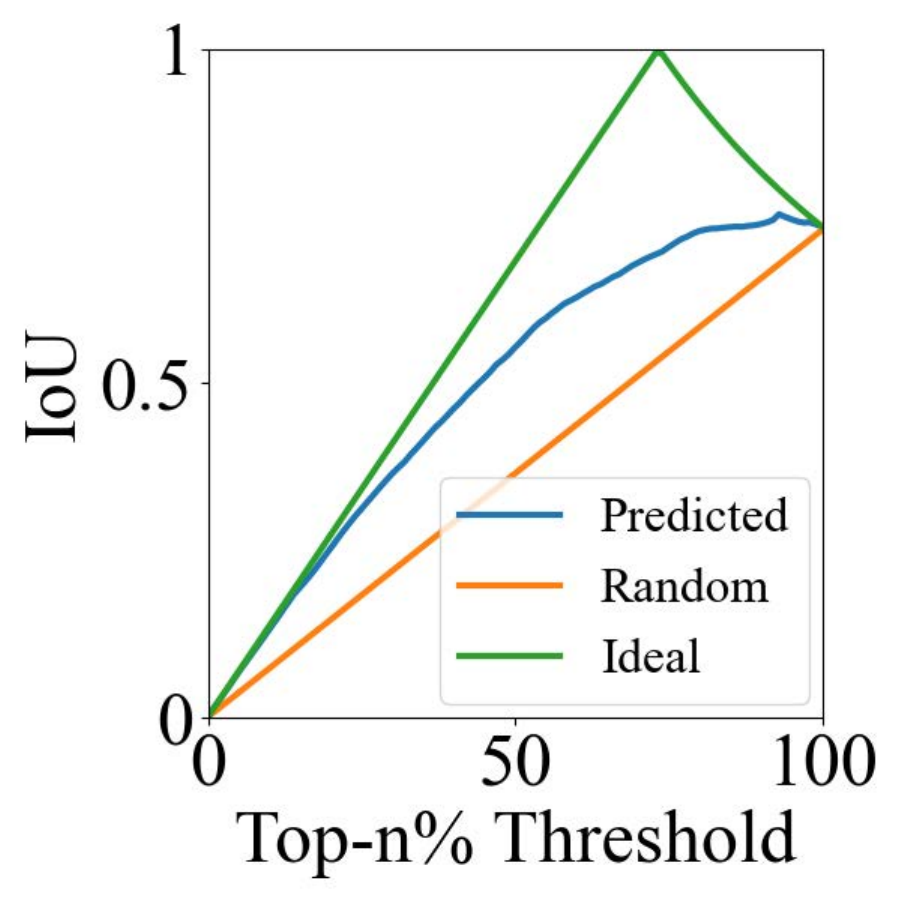}
      			\subcaption{green (NRA: 0.559)}
			\label{fig:4b}
    		\end{minipage}
    	\end{tabular}
\vspace{-8pt}
\caption{Relationship between the top-n\% threshold and segmentation IoU for the predicted concept region (blue). The IoU for the ideal region based on the ground-truth segmentation mask (green), and for a random region (orange) are also shown.}
\label{fig:4}
\end{figure}
\subsection{Evaluation Metrics of Concept Region}
\noindent {\bf Normalized Region Accuracy (NRA).} \indent We propose a novel evaluation metric using segmentation masks. Regions are typically evaluated using the Intersection over Union (IoU) with ground-truth masks. However, saliency maps lack clear boundaries for the target, making evaluation challenging. Visual Explanation Accuracy (VEA)~\cite{VEA, ODAM} is a metric that applies several thresholds to pixel values to define boundaries, allowing comparison between the map and segmentation masks. The relationship between the thresholds and the calculated IoU is plotted, and the area under the curve (AUC) is computed as the evaluation metric. Fig.~\ref{fig:4} presents an example plot for the concept depicted in Fig.~\ref{fig:2}. We also provide plots for cases where concept regions are ideally and randomly obtained with respect to the ground truth. In the random plot, the IoU increases according to the proportion of the mask region relative to the entire image. This illustrates that when the mask region is large, a high AUC can be obtained even for random concept regions. Since the evaluation involves averaging across samples, using a normalized AUC is preferable to ensure independence from the sample distribution. We propose a normalization using the AUC for the ideal plot (${\textrm {AUC}}_{\textrm {high}}$) and that for the random plot (${\textrm {AUC}}_{\textrm {low}}$). This evaluates how close the predicted IoU is to the ideal value. Our proposed metric, Normalized Region Accuracy (NRA), is calculated as follows:
\begin{equation}
\label{eq13}
\textrm {NRA}=\frac{{\textrm {AUC}}-{\textrm {AUC}}_{\textrm {low}}}{{\textrm {AUC}}_{\textrm {high}}-{\textrm {AUC}}_{\textrm {low}}}.
\end{equation}
By using this value, it becomes possible to compare the accuracy across different regions using a unified standard. 
\begin{table}[t]
\footnotesize
\centering
	\begin{tabular}{|>{\centering}m{4.4em}>{\centering}m{7.4em}||>{\centering}m{2.3em}>{\centering}m{2.7em}>{\centering\arraybackslash}m{3.4em}|}
		\hline
		Model & Mehod & EPG Avg. & NRA Avg. & Hit Rate Avg. \\
		\hline
		\multirow{3}{*}{ResNet50} & CLIP-Dissect~\cite{CLIP-dissect} & 0.106 & 0.220 & 0.105 \\
		& WWW~\cite{WWW} & 0.070 & 0.180 & 0.048 \\
		& CE-FAM (Ours) & {\bf 0.124} & {\bf 0.239} & {\bf 0.149} \\
		\hline
		\multirow{3}{*}{ResNet18} & CLIP-Dissect~\cite{CLIP-dissect} & 0.092 & 0.231 & 0.071 \\
		& WWW~\cite{WWW} & 0.079 & 0.216 & 0.038 \\
		& CE-FAM (Ours) & {\bf 0.115} & {\bf 0.236} & {\bf 0.159} \\
		\hline
		\multirow{3}{*}{ViT-B/16} & CLIP-Dissect~\cite{CLIP-dissect} & 0.047 & 0.105 & 0.013 \\
		& WWW~\cite{WWW} & 0.076 & 0.232 & 0.017 \\
		& CE-FAM (Ours) & {\bf 0.138} & {\bf 0.273} & {\bf 0.193} \\
		\hline
        \end{tabular}
\vspace{-8pt}
\caption{Evaluation results when selecting the candidate with the highest association score under the same experimental conditions as in Table~\ref{tab:1}. Only the average across all concept types is reported.}
\label{tab:2}
\end{table}
\vskip.5\baselineskip
\noindent {\bf Energy-based Pointing Game (EPG).} \indent EPG~\cite{ScoreCAM, PG} is a classical benchmark metric for saliency maps in XAI. This metric evaluates the extent to which the pixel values of a saliency map are concentrated on the target object, using the ground-truth region (either a bounding box or a segmentation mask). A higher value indicates that the model relies more on the target object rather than the background, which is considered desirable. The detailed definition is provided in Section~\ref{sup_metric}. However, since EPG does not account for the degree of overlap, it yields high scores even when highly important pixels are merely localized within the target area. Therefore, NRA is more suitable for accurately evaluating the overlap with the target region, and EPG results will be used as a reference in the context of general metrics.

\begin{table}[t]
\footnotesize
\centering
	\begin{tabular}{|>{\centering}m{4.4em}>{\centering}m{7.4em}||>{\centering}m{2.3em}>{\centering}m{2.7em}>{\centering\arraybackslash}m{3.4em}|}
		\hline
		Model & Method & EPG & NRA & Hit Rate\\
		\hline
		\multirow{3}{*}{ResNet50} & CLIP-Dissect~\cite{CLIP-dissect} & 0.614 & 0.588 & 0.662 \\
		& WWW~\cite{WWW} & {\bf 0.625} & 0.595 & 0.674 \\
		& CE-FAM (Ours) & 0.619 & {\bf 0.614} & {\bf 0.744} \\
		\hline
		\multirow{3}{*}{ResNet18} & CLIP-Dissect~\cite{CLIP-dissect} & 0.504 & {\bf 0.563} & 0.260 \\
		& WWW~\cite{WWW} & 0.503 & 0.527 & 0.138 \\
		& CE-FAM (Ours) & {\bf 0.528} & 0.492 & {\bf 0.531} \\
		\hline
		\multirow{3}{*}{ViT-B/16} & CLIP-Dissect & 0.241 & 0.171 & 0.045 \\
		& WWW & 0.437 & 0.448 & 0.038 \\
		& CE-FAM (Ours) & {\bf 0.666} & {\bf 0.671} & {\bf 0.783} \\
		\hline
        \end{tabular}
\vspace{-8pt}
\caption{Evaluation results of concept regions using the ImageNet-S~\cite{ImageNetS} dataset. The evaluation is conducted on all images in the validation split, and the average value is reported.}
\label{tab:3}
\end{table}

\subsection{Quantitative Evaluation of Concept Region}
\label{sec:44}
Table~\ref{tab:1} presents the results of evaluating concept regions using the image and concept sets from Broden~\cite{NetworkDissection}. In concept learning, CLIP is utilized with the Broden concept set $\mathit{T}$. Existing methods perform concept association on the four CNN layers of ResNet under the same settings. Since the frequency of concepts varies depending on their labels, we report the type-wise averages and the average of these values as the final result. We selected the top four region candidates based on their association scores and adopted the one with the highest NRA for the evaluation. While existing methods perform better for concepts related to low-level features such as color, our method outperforms them for other types of concepts as well as in the overall average. These results indicate that higher-level concepts can be effectively represented through the fusion of activation maps. \\
\indent Our method can predict all concepts, therefore all are included in the evaluation. In contrast, existing methods exclude concepts not associated with any activation map as forcing a fixed value for evaluation would make the results dependent on the sample distribution. As a result, associating only high-confidence concepts leads to higher EPG and NRA scores. Therefore, it is also important to evaluate how many concepts are correctly predicted. To address this, we define the Hit Rate, where a concept is considered correctly captured if its NRA exceeds 0.5, and we measure its proportion relative to all samples. The results, shown in Table~\ref{tab:1}, indicate our method outperforms existing methods. \\
\indent Furthermore, Table~\ref{tab:2} presents the evaluation results using the region with the highest association score among the candidates, demonstrating that our method outperforms existing methods in Top-1 matching as well. The detailed results for each type are shown in Appendix Table~\ref{tab:A2}. In Table~\ref{tab:3}, we evaluate ImageNet-S~\cite{ImageNetS} images using ImageNet labels as the concept set. Since all labels are object-related, the evaluation is restricted to the candidate with the highest association score in the final layer. When applied to ViT, representing a concept with a single channel tends to produce noisy results and detailed examples are provided in Fig.~\ref{fig:A3}. In all evaluations, the Hit Rate is higher than that of existing methods, confirming our method's effectiveness.

\begin{figure}[t]
	\centering
	\begin{tabular}{p{0.9\linewidth}}
		\centering
    		\begin{minipage}[t]{0.32\linewidth}
      			\centering
      			\includegraphics[width=1\linewidth]{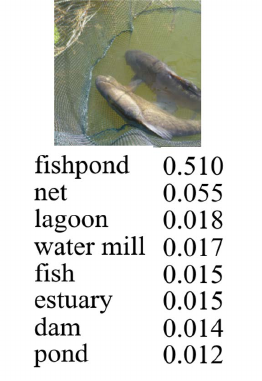}
      			\subcaption{}
			\label{fig:5a}
   		\end{minipage}
		\hspace{10pt}
    		\begin{minipage}[t]{0.48\linewidth}
      			\centering
      			\includegraphics[width=1\linewidth]{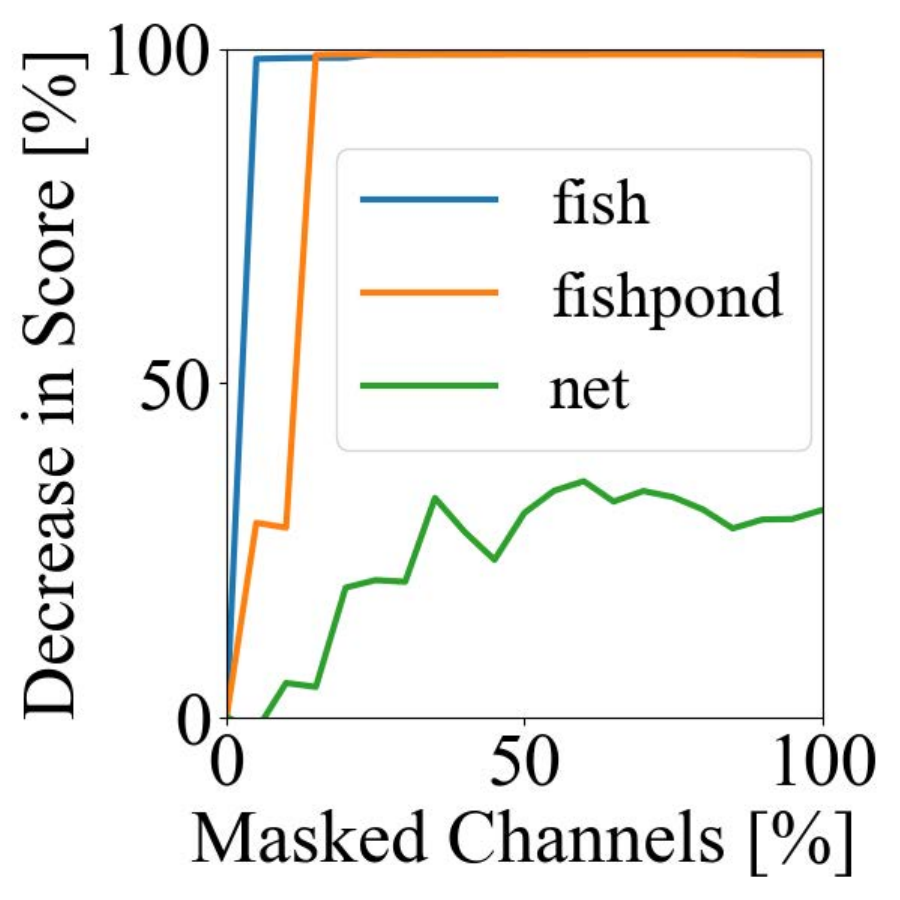}
      			\subcaption{}
			\label{fig:5b}
    		\end{minipage}
    	\end{tabular}
\vspace{-8pt}
\caption{(a) \textit{Tench} image and Broden concept prediction scores using CLIP. (b) Relationship between the decrease in classification score and the proportion of masked channels for certain concepts.}
\label{fig:5}
\end{figure}

\subsection{Evaluation of Concept Contribution}
\label{sec:45}
\begin{figure}[t]
	\centering
	\begin{tabular}{p{0.9\linewidth}}
		\centering
    		\begin{minipage}[t]{1\linewidth}
      			\centering
      			\includegraphics[width=1\linewidth]{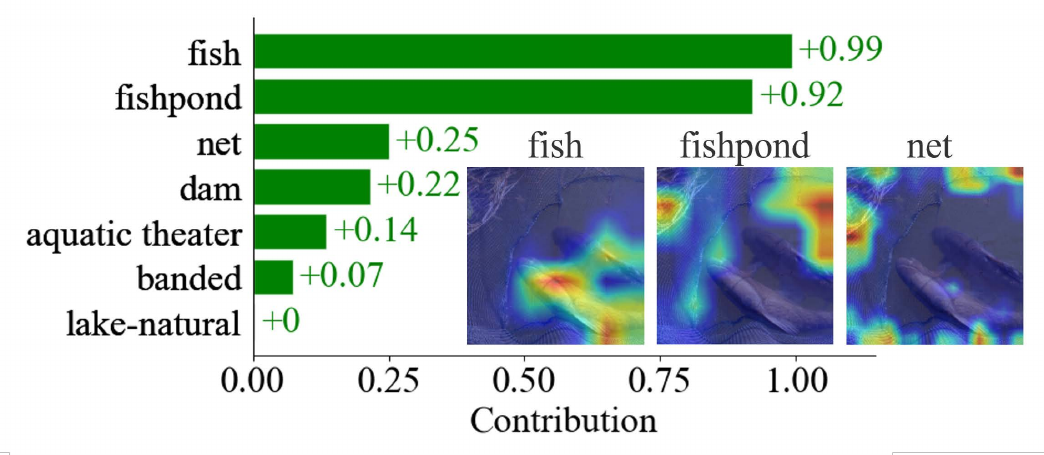}
      			\subcaption{CE-FAM (Ours)}
			\label{fig:6a}
   		\end{minipage} \\
    		\begin{minipage}[t]{1\linewidth}
      			\centering
      			\includegraphics[width=1\linewidth]{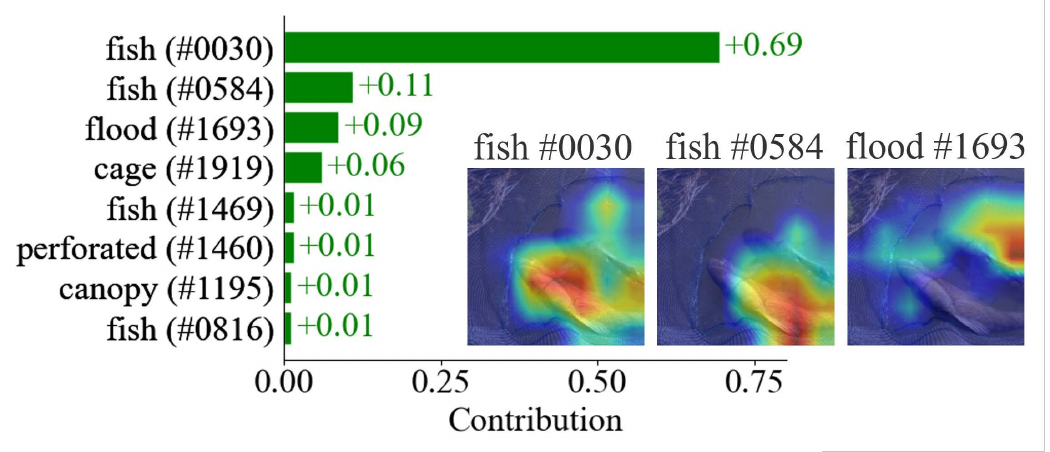}
      			\subcaption{WWW}
			\label{fig:6b}
    		\end{minipage}
    	\end{tabular}
\vspace{-8pt}
\caption{Comparison of concept contributions (image in Fig.~\ref{fig:5a}).}
\label{fig:6}
\end{figure}

\begin{figure}[t]
    \begin{tabular}{p{0.87\linewidth}}
		\centering
        \begin{minipage}[t]{1\linewidth}
      	\centering
      	\includegraphics[width=0.95\linewidth]{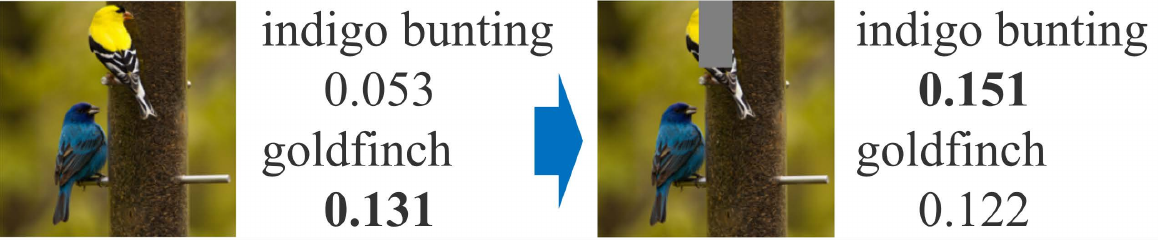}
   		\end{minipage} \\
            \vspace{-1pt}
    	\begin{minipage}[t]{1\linewidth}
      	\centering
      	\includegraphics[width=0.90\linewidth]{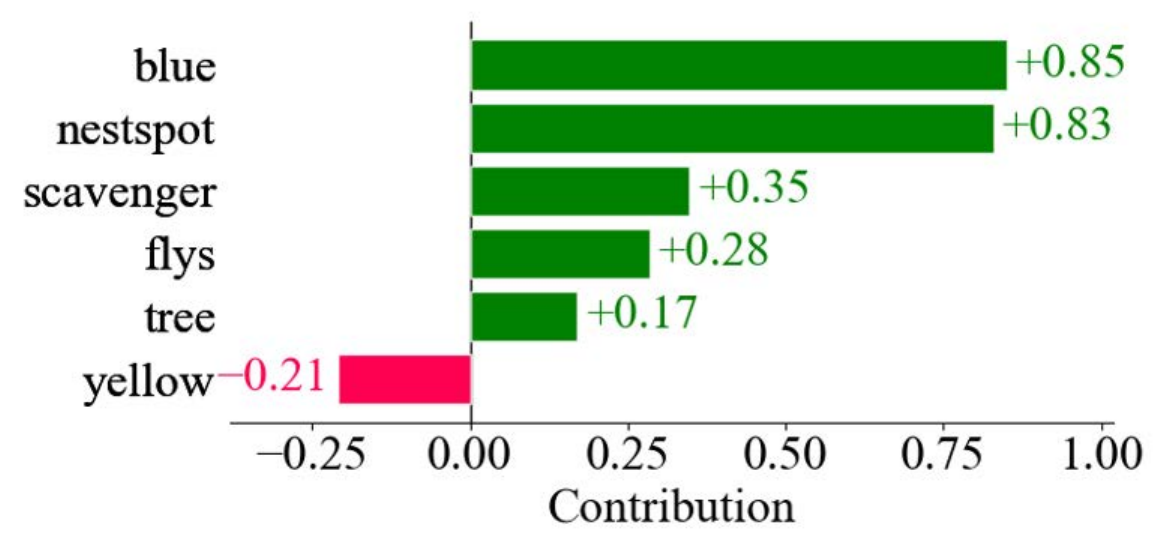}
    	\end{minipage}
    \end{tabular}
\vspace{-15pt}
\caption{Images labeled as \textit{indigo bunting} with prediction scores from ResNet50, and the concept contributions using our method.}
\label{fig:7}
\end{figure}

Since the ground truth of concept contributions is unavailable, we qualitatively evaluate them for ResNet50 on an image shown in Fig.~\ref{fig:5a}. We used the same model as in the experiment in Section~\ref{sec:42}. Fig.~\ref{fig:6} compares the concept contributions, showing the top-ranked concepts along with their associated regions. WWW~\cite{WWW} computes the contributions using an approximation of the Shapley value~\cite{ShapleyValue}. However, this constraint restricts its use to concepts present in \textit{Layer4}. Fig.~\ref{fig:5a} presents the concept prediction scores using CLIP~\cite{CLIP}, where our method successfully predicts similar concepts. In WWW~\cite{WWW}, multiple neurons indicate the concept \textit{fish}, and their corresponding regions are similar. These results suggest that our approach of associating a single concept with multiple neurons is effective. Fig.~\ref{fig:5b} presents a plot of the relationship between masking channels and the decrease in classification scores for calculating the contributions, showing a significant impact from \textit{fish}.\\
\indent We further analyze the factors that led ResNet50~\cite{ResNet} to misclassify an ImageNet image labeled as \textit{indigo bunting} as \textit{goldfinch}. The concept set is based on AwA2~\cite{AWA2}, which contains descriptions of animal attributes. Fig.~\ref{fig:7} presents the concept contributions for predicting \textit{indigo bunting} and corresponding images. We extract and list the concepts with the highest absolute contributions. Notably, \textit{yellow}, a trait of \textit{goldfinch}, has a large negative contribution. By masking the regions corresponding to \textit{yellow} on a patch basis, we observed a score reversal, leading to correct classification.

\subsection{Discussion}
\label{sec:discussion}
\noindent {\bf Ablation study.} \indent To analyze the key factors in concept learning, we conducted training and evaluation under various conditions. The baseline evaluation setting is equivalent to the one using ResNet50~\cite{ResNet} in Table~\ref{tab:1}. The experimental conditions include the choice of the VLM, the use of Similarity Loss $\mathcal{L}_{\textrm {sim}}$, and the utilization of multiple layers. When multiple layers are not used, $\mathbf{z}^L$ is used for learning instead of $\mathbf{z}_{\textrm{cat}}$. According to the results shown in Table~\ref{tab:4}, the use of $\mathcal{L}_{\textrm {sim}}$ and multiple layers contributes to improving the accuracy of concept regions. Additionally, SigLIP~\cite{SigLIP} outperforms CLIP, demonstrating its effectiveness in learning concept representations. When $\mathcal{L}_{\textrm {sim}}$ is not used, training is conducted without a predefined concept set, yet it still achieves higher performance than the existing methods in Table~\ref{tab:1}. By mimicking VLM's image embeddings, our method is also effective in a zero-shot setting with respect to concept labels, providing an advantage over existing methods that can only be applied to predefined concepts.
\vskip.2\baselineskip
\noindent {\bf Effect of model training.} \indent Fig.~\ref{tab:A2} in the appendix shows the Hit Rate of CE-FAM improves with training epochs, outperforming the existing methods in Table~\ref{tab:1} within just a few epochs. Since training is performed once and can be parallelized, we consider the required time acceptable.
\vskip.2\baselineskip
\noindent {\bf Effect of normalization.} \indent Table~\ref{tab:A4} presents unnormalized NRA (i.e., AUC), but the differences between methods are less pronounced. We report average ${\textrm {AUC}}_{\textrm {high}}$ and ${\textrm {AUC}}_{\textrm {low}}$ across all images, though they vary by category and affect the value range. If the aim is to locate results between upper and lower bounds, normalization is reasonable.

\subsection{Limitation}
Our proposed method, which establishes a one-to-one correspondence between concept labels, regions, and contributions, offers a more intuitive framework for humans. However, it faces two challenges. The first challenge is that the expressiveness of concepts is constrained by the performance of the VLM. As shown in Fig.~\ref{fig:5a}, CLIP tends to be influenced by prominent features in an image, making it difficult to predict finer-grained concepts. Additional strategies are needed to accurately capture detailed concepts. The second challenge is the appropriate selection of the concept set. If the set is excessively large, the method becomes more susceptible to noise from irrelevant concepts. It is crucial to select a concept set that is both necessary and sufficient. While we demonstrated the generalizability of our method across various major settings, its broader applicability remains unexplored. Moreover, although concept contributions are intuitively measured by their impact on classification scores, establishing a quantitative metric is difficult, hindering effective validation. These challenges are shared by existing methods and remain open problems. Future work includes extending evaluations to other settings and optimizing our framework, including the model training.

\begin{table}[t]
\centering
\small
\begin{tabular}{|>{\centering}m{2.8em}>{\centering}m{4.0em}>{\centering}m{2.8em}||>{\centering}m{2.4em}>{\centering}m{2.4em}>{\centering\arraybackslash}m{2.4em}|}
	\hline
	VLM Model & Similarity Loss & Multi Layer& EPG Avg. & NRA Avg. & Hit Avg. \\
	\hline
	CLIP & & \checkmark & 0.152 & 0.338 & 0.209 \\
	SigLIP & & \checkmark & 0.151 & 0.383 & 0.283 \\
	CLIP & \checkmark & & 0.156 & 0.348 & 0.234 \\
	SigLIP & \checkmark & & 0.156 & 0.374 & 0.274 \\
	CLIP & \checkmark & \checkmark & 0.154 & 0.361 & 0.247 \\
	SigLIP & \checkmark & \checkmark & {\bf 0.157} & {\bf 0.388} & {\bf 0.295} \\
	\hline
\end{tabular}
\vspace{-10pt}
\caption{Concept region evaluation under different conditions for concept learning, showing the average across all concept types.}
\label{tab:4}
\end{table}
\section{Conclusions}
\label{sec:5_conclusion}
We proposed a novel concept-based explanation method that identifies the region and contribution associated with each concept. We also introduced a new metric to assess the accuracy of concept regions, and evaluations under various conditions demonstrated the superiority of our method. Our framework is versatile and can be extended to various models, and performance improvements can be expected as VLMs advance. As a future direction, exploring more robust approaches to concept learning would be beneficial.
{
    \small
    \bibliographystyle{ieeenat_fullname}
    \bibliography{main}
}
\clearpage
\setcounter{page}{1}
\setcounter{section}{0}
\setcounter{figure}{0}
\setcounter{table}{0}
\setcounter{equation}{0}
\maketitlesupplementary
\setlength\textfloatsep{3pt}
\setlength\floatsep{3pt}
\renewcommand{\thesection}{\Alph{section}}
\renewcommand{\thefigure}{A\arabic{figure}}
\renewcommand{\thetable}{A\arabic{table}}
\renewcommand{\theequation}{A\arabic{equation}}
\renewcommand{\thepage}{A\arabic{page}}

\section{Experimental Details}
\subsection{Dataset Details}
\noindent {\bf Broden dataset.} \indent This dataset~\cite{NetworkDissection} is widely used in research on concept-based explainable AI~\cite{NetworkDissection, CLIP-dissect, WWW}. The images have a resolution of $224 \times 224$ pixels, and are accompanied by segmentation masks corresponding to the concepts present in the images. The concept labels are divided into multiple categories, with examples provided in Table ~\ref{tab:A1}.
\vskip.5\baselineskip
\noindent {\bf ImageNet dataset.} \indent This dataset~\cite{ImageNet} is widely used in various studies on image classification. The images have varying resolutions but are resized to $224 \times 224$ pixels when input to the model. The dataset contains 1,000 labels related to objects. The training set contains over one million images, while the validation set contains 50,000 images. We used the training set for concept learning and the validation set for evaluation. Moreover, ImageNet-S~\cite{ImageNetS} extends ImageNet by providing segmentation masks for target objects.
\vskip.5\baselineskip
\noindent {\bf AwA2 dataset.} \indent This dataset~\cite{AWA2} is commonly used for zero-shot learning tasks. It contains over 30,000 animal images with 50 animal-related classes and 85 attribute labels. These attributes include information about color, animal characteristics, and the surrounding environment. Since they effectively represent concepts related to animals, we used the attribute labels as the concept set in our evaluation shown in Fig.~\ref{fig:7}.

\subsection{Model Details}
\noindent {\bf Image Classifier and VLM.} \indent ResNet50, ResNet18~\cite{ResNet}, and ViT-B/16~\cite{ViT} models used in this study are ImageNet-pretrained models available in PyTorch. We used the same models for the implementation of both our method and the baseline methods. For the Vision and Language Models (VLMs), we used the publicly available pretrained models of CLIP~\cite{CLIP} and SigLIP~\cite{SigLIP}.
\vskip.5\baselineskip
\noindent {\bf Translator function for concept learning.} \indent For the translator function $h$ described in Eq.~\ref{eq6}, we employed a simple Multi-Layer Perceptron (MLP). In ResNet50~\cite{ResNet}, the embedding vectors $\mathbf{z}^1$, $\mathbf{z}^2$, $\mathbf{z}^3$, and $\mathbf{z}^4$, have dimensions of 256, 512, 1024, and 2048, respectively, resulting in a concatenated vector $\mathbf{z}_{\textrm {cat}}$ of dimension 3840. The MLP consists of three layers with the following input and output dimensions: (3840, 2048), (2048, 1024), and (1024, 512). The output dimension of 512 matches that of CLIP's image embeddings. Since the VLM's image embeddings include negative values, we use the \textit{tanh} activation function to account for the influence of negative values during backpropagation when computing gradients.
\vskip.5\baselineskip
\noindent {\bf Implementation of baseline methods.} \indent For the implementations of CLIP-Dissect~\cite{CLIP-dissect} and WWW~\cite{WWW}, we used publicly available source code. Since both implementations support the labels from the Broden~\cite{NetworkDissection} and ImageNet~\cite{ImageNet} datasets, we leveraged them in our experiments. All implementation parameters were kept at their default settings.

\begin{figure}[t]
\centering
    \begin{tabular}{p{1\linewidth}}
    \centering
        \begin{minipage}[t]{0.73\linewidth}
        \centering
            \begin{minipage}[t]{1\linewidth}
                \centering
                \includegraphics[width=1\linewidth]{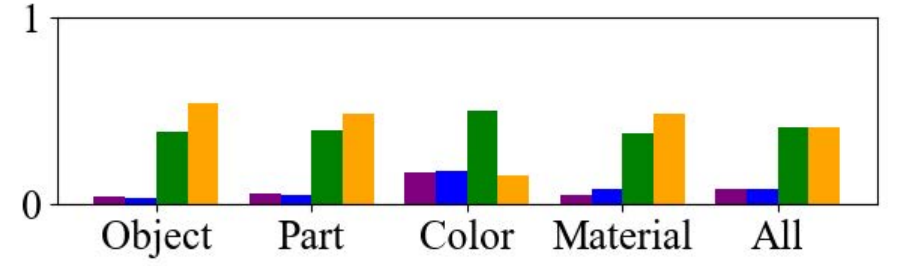}
                \vspace{-15pt}
                \subcaption{CE-FAM (Ours)}
            \end{minipage} \\ 
            \begin{minipage}[t]{1\linewidth}
                \centering
                \includegraphics[width=1\linewidth]{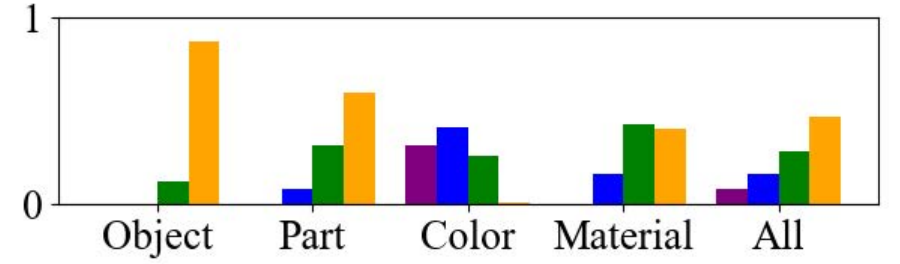}
                \vspace{-15pt}
                \subcaption{CLIP-Dissect~\cite{CLIP-dissect}}
            \end{minipage} \\ 
            \begin{minipage}[t]{1\linewidth}
                \centering
                \includegraphics[width=1\linewidth]{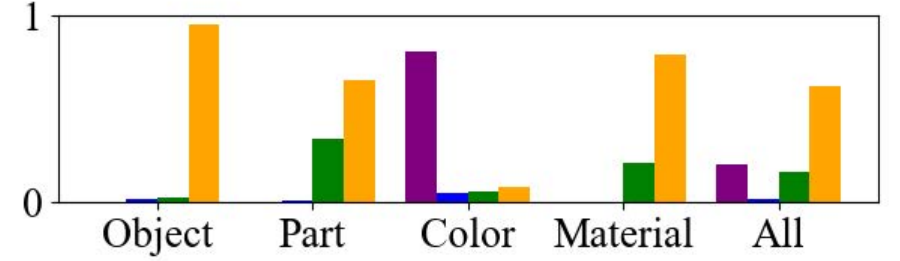}
                \vspace{-15pt}
                \subcaption{WWW~\cite{WWW}}
            \end{minipage} \\
        \end{minipage}
        \begin{minipage}[c]{0.23\linewidth}
            \centering
            \includegraphics[width=1\linewidth]{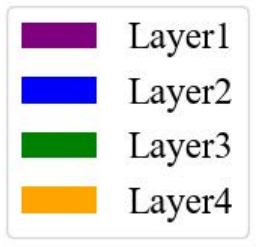}
            \vfill
        \end{minipage}
    \end{tabular}
\vspace{-10pt}
\caption{Proportion of layers from which the candidate with the highest NRA was selected in the experiment using ResNet50~\cite{ResNet} in Table~\ref{tab:1}.}
\label{fig:A1}
\end{figure}

\subsection{Metric Details}
\label{sup_metric}
\noindent {\bf Intersection over Union (IoU).} \indent We explain the calculation of IoU between the concept region and the segmentation mask. Here, we define the concept region map as $\mathcal{R}$ and the segmentation mask as $\mathcal{M}$. In our metric, we utilize a threshold $T_n$ that separates the top $n$\% pixel values of $\mathcal{R}$ to define the boundary of the concept region. The IoU for each $T_n$ is given by the following equation.
\begin{equation}
\text{IoU}_{n} = \frac{|(\mathcal{R} \geq T_n) \cap \mathcal{M}|}{|(\mathcal{R} \geq T_n) \cup \mathcal{M}|}.
\end{equation}
The area under the curve (AUC), obtained by plotting the relationship between the threshold $T_n$ and $\text{IoU}_{n}$, is used to compute our metric, Normalized Region Accuracy (NRA).
\vskip.5\baselineskip
\noindent {\bf Energy-based Pointing Game (EPG).} \indent The evaluation metric EPG~\cite{ScoreCAM} in the main paper is defined as follows.
\begin{equation}
\textrm {EPG} = \frac{\sum{}\mathcal{R}_{(i,j)\in \mathcal{M}}}{\sum{}\mathcal{R}_{(i,j)\in \mathcal{M}}+\sum{}\mathcal{R}_{(i,j)\notin \mathcal{M}}}.
\end{equation}
$\mathcal{R}_{(i,j)}$ denotes the pixel value at $(i, j)$ in the concept region map. \textrm {EPG} calculates the proportion of the total sum of $\mathcal{R}_{(i,j)}$ for the pixels $(i,j)$ in the segmentation mask $\mathcal{M}$.

\begin{table*}[t]
\small
\centering
	\begin{tabular}{|>{\centering}m{7.4em}||>{\centering}m{10.0em}>{\centering}m{10.0em}>{\centering}m{10.0em}>{\centering\arraybackslash}m{10.5em}|}
		\hline
		Category & Object & Part & Color & Material \\
        \hline
        Number of labels & 584 & 234 & 11 & 32 \\
        \hline
        Example & wall, sky, floor, tree, $\dots$ & head, leg, torso, arm, $\dots$ & black, grey, white, $\dots$ & wood, painted, fabric, $\dots$ \\
        \hline
        frequency & 271,047 & 132,414 & 651,776 & 54,511 \\
        \hline
        Template & ``a photo of a \{ \}'' & ``a photo of a \{ \}'' &  ``a photo of a \{ \} object'' & ``a photo of an object made of \{ \}''\\
		\hline
        \end{tabular}
\vspace{-8pt}
\caption{Details of the concept labels defined in the Broden~\cite{NetworkDissection} dataset. Other categories, such as scene and texture, exist but are excluded from the evaluation as segmentation masks are not provided. ``Template'' refers to the textual format of concept labels used in the proposed method and is inspired by those used in CLIP~\cite{CLIP} training.}
\label{tab:A1}
\end{table*}

\begin{table*}[t]
\footnotesize
\centering
	\begin{tabular}{|>{\centering}m{3.0em}>{\centering}m{5.8em}||>{\centering}m{1.8em}>{\centering}m{1.8em}>{\centering}m{1.8em}>{\centering}m{1.8em}>{\centering}m{1.8em}|>{\centering}m{1.8em}>{\centering}m{1.8em}>{\centering}m{1.8em}>{\centering}m{1.8em}>{\centering}m{1.8em}|>{\centering}m{1.8em}>{\centering}m{1.8em}>{\centering}m{1.8em}>{\centering}m{1.8em}>{\centering\arraybackslash}m{1.8em}|}
		\hline
		& & \multicolumn{5}{c|}{EPG~\cite{ScoreCAM}} & \multicolumn{5}{c|}{NRA} & \multicolumn{5}{c|}{Hit Rate}\\
		\cline{3-17}
		Model & Method & Object & Part & Color & Mate-rial & Avg. & Object & Part & Color & Mate-rial & Avg. & Object & Part & Color & Mate-rial & Avg. \\
		\hline
		\multirow{3}{*}{ResNet50} & CLIP-Dissect & 0.161 & 0.033 & {\bf 0.132} & 0.097 & 0.106 & 0.253 & 0.169 & {\bf 0.279} & 0.178 & 0.220 & 0.147 & 0.050 & {\bf 0.154} & 0.070 & 0.105\\
		& WWW & 0.145 & 0.038 & 0.031 & 0.068 & 0.070 & 0.256 & 0.153 & 0.145 & 0.165 & 0.180 & 0.110 & 0.025 & 0.025 & 0.032 & 0.048\\
		& CE-FAM & {\bf 0.210} & {\bf0.049} & 0.127 & {\bf 0.110} & {\bf 0.124} & {\bf 0.342} & {\bf 0.194} & 0.179 & {\bf 0.241} & {\bf 0.239} & {\bf 0.299} & {\bf 0.110} & 0.044 & {\bf 0.145} & {\bf 0.149}\\
		\hline
		\multirow{3}{*}{ResNet18} & CLIP-Dissect & 0.152 & 0.038 & 0.100 & 0.079 & 0.092 & 0.280 & {\bf 0.199} & {\bf 0.270} & 0.175 & 0.231 & 0.090 & 0.031 & {\bf 0.113} & 0.052 & 0.071\\
		& WWW & 0.113 & 0.033 & 0.112 & 0.058 & 0.079 & 0.233 & 0.168 & 0.234 & 0.231 & 0.216 & 0.031 & 0.010 & 0.098 & 0.014 & 0.038\\
		& CE-FAM & {\bf 0.192} & {\bf 0.043} & {\bf 0.115} & {\bf 0.110} & {\bf 0.115} & {\bf 0.343} & 0.190 & 0.172 & {\bf 0.240} & {\bf 0.236} & {\bf 0.308} & {\bf 0.108} & 0.049 & {\bf 0.171} & {\bf 0.159}\\
		\hline
		\multirow{3}{*}{ViT-B/16} & CLIP-Dissect & 0.095 & 0.022 & 0.000 & 0.071 & 0.047 & 0.134 & 0.108 & 0.000 & 0.177 & 0.105 & 0.013 & 0.005 & 0.000 & 0.033 & 0.013\\
		& WWW & 0.070 & 0.033 & 0.122 & 0.079 & 0.076 & 0.233 & 0.206 & 0.170 & {\bf 0.318} & 0.232 & 0.018 & 0.004 & 0.039 & 0.005 & 0.017\\
		& CE-FAM & {\bf 0.240} & {\bf 0.063} & {\bf 0.127} & {\bf 0.122} & {\bf 0.138} & {\bf 0.401} & {\bf 0.253} & {\bf 0.183} & 0.253 & {\bf 0.273} & {\bf 0.385} & {\bf 0.169} & {\bf 0.058} & {\bf 0.161} & {\bf 0.193}\\
		\hline
        \end{tabular}
\vspace{-8pt}
\caption{Detailed results for each concept type corresponding to the experimental results in Table~\ref{tab:2}.}
\label{tab:A2}
\end{table*}

\begin{table*}[t]
\footnotesize
\centering
	\begin{tabular}{|>{\centering}m{3.4em}>{\centering}m{5.8em}||>{\centering}m{1.8em}>{\centering}m{1.8em}>{\centering}m{1.8em}>{\centering}m{1.8em}>{\centering}m{1.7em}|>{\centering}m{1.8em}>{\centering}m{1.8em}>{\centering}m{1.8em}>{\centering}m{1.8em}>{\centering}m{1.7em}|>{\centering}m{1.8em}>{\centering}m{1.8em}>{\centering}m{1.8em}>{\centering}m{1.8em}>{\centering\arraybackslash}m{1.7em}|}
		\hline
		& & \multicolumn{5}{c|}{EPG~\cite{ScoreCAM}} & \multicolumn{5}{c|}{NRA} & \multicolumn{5}{c|}{Hit Rate}\\
		\cline{3-17}
		Model & Method & Object & Part & Color & Mate-rial & Avg. & Object & Part & Color & Mate-rial & Avg. & Object & Part & Color & Mate-rial & Avg. \\
		\hline
		\multirow{3}{*}{\shortstack{ResNet18 \\ (Places365)}} & CLIP-Dissect & 0.195 & 0.035 & 0.143 & 0.098 & 0.118 & 0.340 & 0.200 & {\bf 0.431} & 0.221 & 0.298 & 0.125 & 0.017 & 0.260 & 0.054 & 0.114\\
		& WWW & 0.201 & 0.051 & 0.161 & 0.128 & 0.136 & 0.397 & 0.222 & 0.400 & {\bf 0.496} & {\bf 0.379} & 0.077 & 0.012 & {\bf 0.264} & 0.011 & 0.091 \\
		& CE-FAM & {\bf 0.236} & {\bf 0.054} & {\bf 0.162} & {\bf 0.158} & {\bf 0.152} & {\bf 0.411} & {\bf 0.269} & 0.315 & 0.358 & 0.338 & {\bf 0.351} & {\bf 0.118} & 0.165 & {\bf 0.248} & {\bf 0.221}\\
		\hline
        \end{tabular}
\vspace{-10pt}
\caption{Evaluation results for concept regions using all images from the Broden~\cite{NetworkDissection} dataset. The highest NRA among the top 4 scoring candidates is evaluated. The average value for each concept type is reported, and "Avg." denotes the mean of these per-type averages. A ResNet18~\cite{ResNet} pretrained on Places365~\cite{Places365} is used as the image classifier.}
\label{tab:A3}
\end{table*}

\section{Analysis of Experimental Results}
\subsection{Concept Region for Broden dataset}
\begin{figure}[t]
\centering
    \centering
    \includegraphics[width=0.75\linewidth]{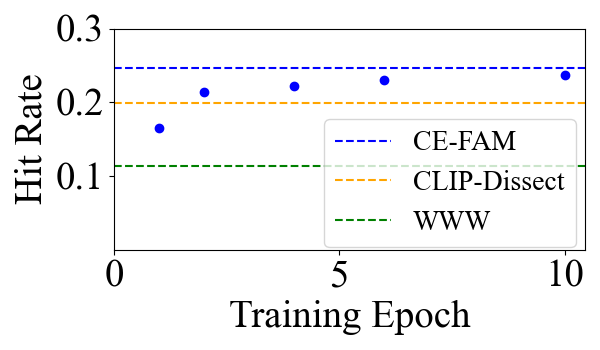}
\vspace{-10pt}
\caption{Improvement in the Hit Rate over training epochs under the same conditions as in Table~\ref{tab:1}. The blue plots indicate the results of the proposed method, and the dotted lines show the results from Table~\ref{tab:1} for reference.}
\label{fig:A2}
\end{figure}

\noindent {\bf Baseline evaluation.} \indent In the experiment presented in Table~\ref{tab:1}, the top four candidates with the highest association scores were extracted, and the one with the highest NRA among them was used for evaluation. This approach was taken because, as shown in Fig.~\ref{fig:2}, the neuron with the highest association score does not necessarily yield the best result in existing methods. Since the proposed method can generate four candidates from \textit{Layer1} to \textit{Layer4}, the evaluation was conducted using four candidates in all methods. Furthermore, Fig.~\ref{fig:A1} presents the results indicating from which layer the candidate with the highest NRA was selected. These results correspond to the experiment conducted with ResNet50~\cite{ResNet} in Table~\ref{tab:1}. For concept types such as \textit{Color}, effectively extracting information from lower layers is crucial. Additionally, for other concept types, it is important to extract information not only from \textit{Layer4} but also from \textit{Layer3}. In the evaluation results presented in Table~\ref{tab:2}, detailed results for each concept type were omitted. We provide these results in Table~\ref{tab:A2}. The proposed method performs worse than existing methods for the \textit{Color} type but outperforms them for other concept types. For concepts based on lower-level features, such as \textit{Color}, a small number of channels in the activation maps are sufficient for representation, and taking a weighted sum over multiple channels may introduce noise. If this noise can be reduced during concept learning, the proposed method is expected to achieve higher accuracy even for concepts based on lower-level features.
\vskip.5\baselineskip
\noindent {\bf Extended evaluation.} \indent To verify the effectiveness of our method, we conducted an extended evaluation beyond the baseline. In the baseline setting, we used ResNet~\cite{ResNet} models pre-trained on ImageNet~\cite{ImageNet}, as commonly adopted in existing methods. Since prior works~\cite{CLIP-dissect, WWW} also evaluated a ResNet18 pre-trained on the Places365~\cite{Places365} dataset, which was publicly available in their implementation, we included it in our evaluation. The results are shown in Table~\ref{tab:A3}, using CLIP~\cite{CLIP} as the VLM and the Broden~\cite{NetworkDissection} dataset for evaluation, under the same experimental conditions as in Table~\ref{tab:1}. This model also exhibits trends consistent with those observed in other experiments.\\
\indent To examine the convergence of the training module, Fig.~\ref{fig:A2} shows the change in the Hit Rate across training epochs under the same conditions as ResNet50 in Table~\ref{tab:1}. We observe that even after a few epochs, the results already surpass those of the existing methods in Table~\ref{tab:1}. While longer training improves the accuracy of concept learning, it involves a trade-off with training cost. Similar to knowledge distillation, mimicking VLM embeddings requires only diverse images. However, for better accuracy on specific concepts, domain-relevant images are more effective, as with concept matching in existing methods~\cite{CLIP-dissect, WWW}.\\
\indent To validate the effectiveness of the evaluation metric, we also present results without normalization. The proposed NRA is designed to provide fair evaluation across different concept labels by applying normalization. Table~\ref{tab:A4} shows the results when only AUC is used without normalization. The experimental conditions follow those of ResNet50 in Table~\ref{tab:1}, and the same translator function model is used for evaluation. As shown by the average values of ${\textrm {AUC}}_{\textrm {high}}$ and ${\textrm {AUC}}_{\textrm {low}}$ for each category, the maximum and minimum AUC values vary significantly across categories. These differences in value ranges can introduce bias in the evaluation results, making the evaluation sensitive to the distribution of sample sizes. By applying the proposed normalization, the influence of concept region size distribution can be reduced, enabling fairer evaluation.

\begin{figure*}[t]
\centering
    \begin{tabular}{p{1\linewidth}}
    \centering
        \begin{minipage}[t]{0.48\linewidth}
            \centering
            \begin{minipage}[t]{0.23\linewidth}
                \centering
                \subcaption*{rapeseed}
                \includegraphics[width=0.8\linewidth]{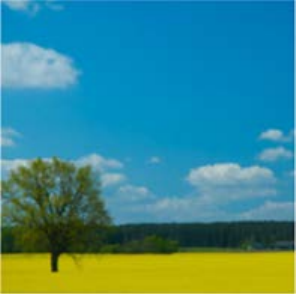}
            \end{minipage}
            \hspace{-8pt}
            \begin{minipage}[t]{0.23\linewidth}
                \centering
                \subcaption*{CE-FAM}
                \includegraphics[width=0.8\linewidth]{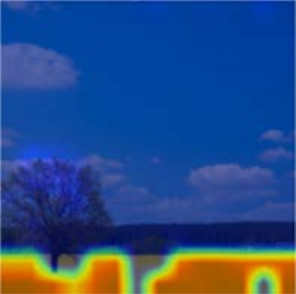}
            \end{minipage}
            \hspace{-8pt}
            \begin{minipage}[t]{0.23\linewidth}
                \centering
                \subcaption*{CLIP-Dissect}
                \includegraphics[width=0.8\linewidth]{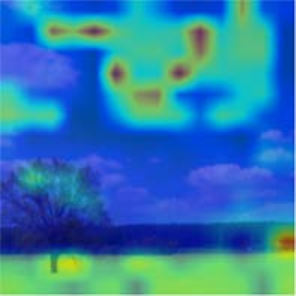}
            \end{minipage}
            \hspace{-8pt}
            \begin{minipage}[t]{0.23\linewidth}
                \centering
                \subcaption*{WWW}
                \includegraphics[width=0.8\linewidth]{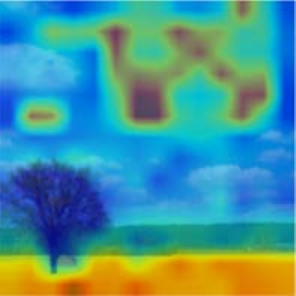}
            \end{minipage} \\
            \begin{minipage}[t]{0.23\linewidth}
                \centering
                \subcaption*{limousine}
                \includegraphics[width=0.8\linewidth]{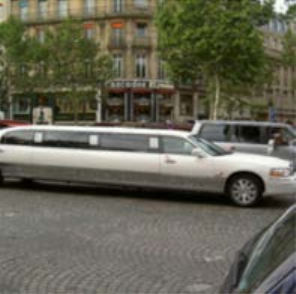}
            \end{minipage}
            \hspace{-8pt}
            \begin{minipage}[t]{0.23\linewidth}
                \centering
                \subcaption*{CE-FAM}
                \includegraphics[width=0.8\linewidth]{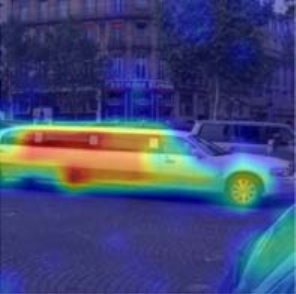}
            \end{minipage}
            \hspace{-8pt}
            \begin{minipage}[t]{0.23\linewidth}
                \centering
                \subcaption*{CLIP-Dissect}
                \includegraphics[width=0.8\linewidth]{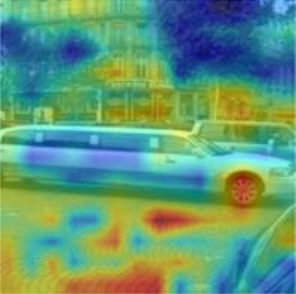}
            \end{minipage}
            \hspace{-8pt}
            \begin{minipage}[t]{0.23\linewidth}
                \centering
                \subcaption*{WWW}
                \includegraphics[width=0.8\linewidth]{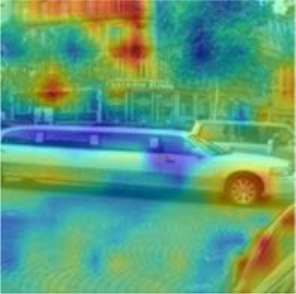}
            \end{minipage} \\
            \begin{minipage}[t]{0.23\linewidth}
                \centering
                \subcaption*{monarch}
                \includegraphics[width=0.8\linewidth]{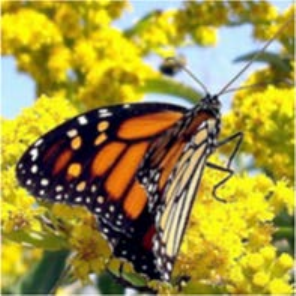}
            \end{minipage}
            \hspace{-8pt}
            \begin{minipage}[t]{0.23\linewidth}
                \centering
                \subcaption*{CE-FAM}
                \includegraphics[width=0.8\linewidth]{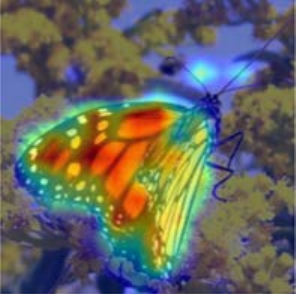}
            \end{minipage}
            \hspace{-8pt}
            \begin{minipage}[t]{0.23\linewidth}
                \centering
                \subcaption*{CLIP-Dissect}
                \includegraphics[width=0.8\linewidth]{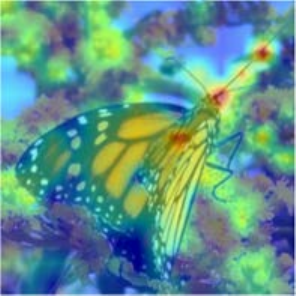}
            \end{minipage}
            \hspace{-8pt}
            \begin{minipage}[t]{0.23\linewidth}
                \centering
                \subcaption*{WWW}
                \includegraphics[width=0.8\linewidth]{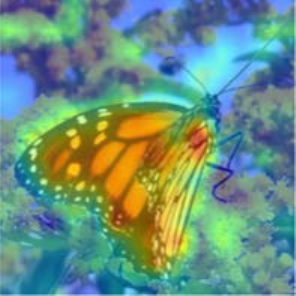}
            \end{minipage} \\
        \subcaption{ImageNet~\cite{ImageNet}}
        \label{fig:A2a}
        \end{minipage}
        \begin{minipage}[t]{0.48\linewidth}
            \centering
            \begin{minipage}[t]{0.23\linewidth}
                \centering
                \subcaption*{cage}
                \includegraphics[width=0.8\linewidth]{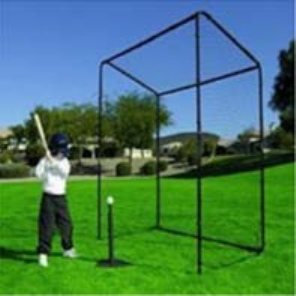}
            \end{minipage}
            \hspace{-8pt}
            \begin{minipage}[t]{0.23\linewidth}
                \centering
                \subcaption*{CE-FAM}
                \includegraphics[width=0.8\linewidth]{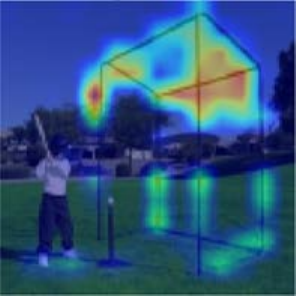}
            \end{minipage}
            \hspace{-8pt}
            \begin{minipage}[t]{0.23\linewidth}
                \centering
                \subcaption*{CLIP-Dissect}
                \includegraphics[width=0.8\linewidth]{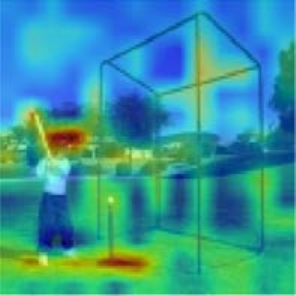}
            \end{minipage}
            \hspace{-8pt}
            \begin{minipage}[t]{0.23\linewidth}
                \centering
                \subcaption*{WWW}
                \includegraphics[width=0.8\linewidth]{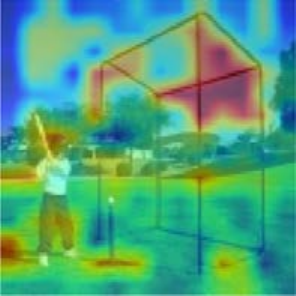}
            \end{minipage} \\
            \begin{minipage}[t]{0.23\linewidth}
                \centering
                \subcaption*{banner}
                \includegraphics[width=0.8\linewidth]{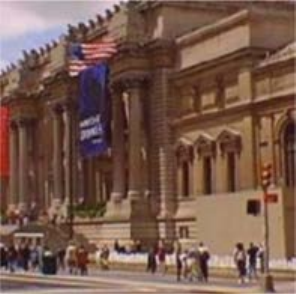}
            \end{minipage}
            \hspace{-8pt}
            \begin{minipage}[t]{0.23\linewidth}
                \centering
                \subcaption*{CE-FAM}
                \includegraphics[width=0.8\linewidth]{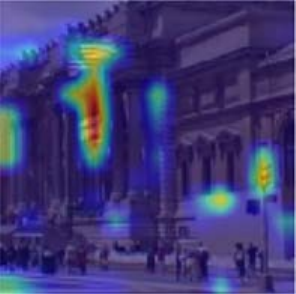}
            \end{minipage}
            \hspace{-8pt}
            \begin{minipage}[t]{0.23\linewidth}
                \centering
                \subcaption*{CLIP-Dissect}
                \includegraphics[width=0.8\linewidth]{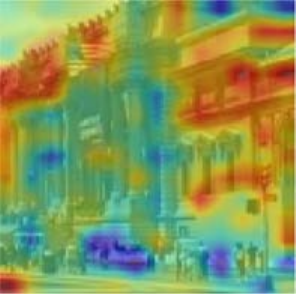}
            \end{minipage}
            \hspace{-8pt}
            \begin{minipage}[t]{0.23\linewidth}
                \centering
                \subcaption*{WWW}
                \includegraphics[width=0.8\linewidth]{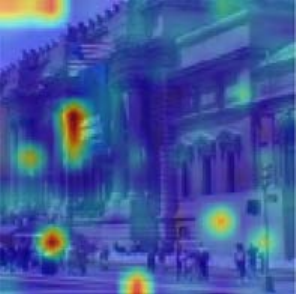}
            \end{minipage} \\
            \begin{minipage}[t]{0.23\linewidth}
                \centering
                \subcaption*{curtain}
                \includegraphics[width=0.8\linewidth]{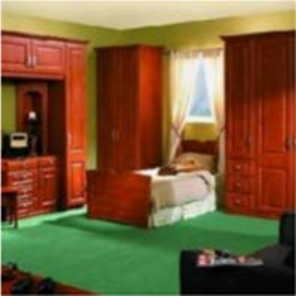}
            \end{minipage}
            \hspace{-8pt}
            \begin{minipage}[t]{0.23\linewidth}
                \centering
                \subcaption*{CE-FAM}
                \includegraphics[width=0.8\linewidth]{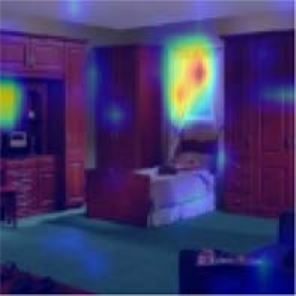}
            \end{minipage}
            \hspace{-8pt}
            \begin{minipage}[t]{0.23\linewidth}
                \centering
                \subcaption*{CLIP-Dissect}
                \includegraphics[width=0.8\linewidth]{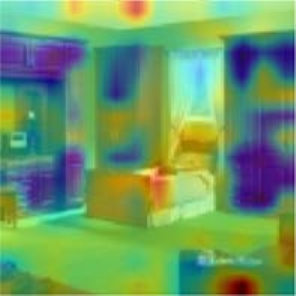}
            \end{minipage}
            \hspace{-8pt}
            \begin{minipage}[t]{0.23\linewidth}
                \centering
                \subcaption*{WWW}
                \includegraphics[width=0.8\linewidth]{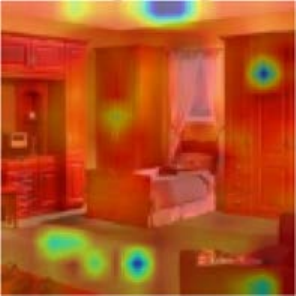}
            \end{minipage} \\
        \subcaption{Broden~\cite{NetworkDissection}}
        \label{fig:A2b}
	    \end{minipage} 
    \end{tabular}
\vspace{-10pt}
\caption{Comparison of concept regions obtained when ViT~\cite{ViT} is used as the target image classifier. (a) shows the results using ImageNet~\cite{ImageNet} as the image and concept set, while (b) shows the results using Broden~\cite{NetworkDissection}.}
\label{fig:A3}
\end{figure*}

\subsection{Concept Region for ImageNet-S dataset}
Since ImageNet~\cite{ImageNet} contains labels for objects, the evaluation in Table~\ref{tab:3} is conducted using only the candidate with the highest association score from the final layer (\textit{Layer4}). This approach is based on prior findings in WWW~\cite{WWW}, which report that concepts associated with the classification classes are strongly observed in the final layer. Therefore, concept learning in our method utilizes only $\mathbf{z}^L$ rather than the concatenated vector $\mathbf{z}_{\textrm {cat}}$ from all layers in this evaluation. As shown in Table~\ref{tab:3}, all methods achieve high accuracy, suggesting that object-related information is concentrated in the final layer. Although our method demonstrates a significant advantage in the Hit Rate, the differences in EPG~\cite{ScoreCAM} and NRA are relatively small. This is because existing methods exclude concepts that do not correspond to any neuron in these evaluations. Consequently, the evaluations are conducted on neurons with relatively high confidence. In contrast, Hit Rate is calculated as the proportion over all samples without these exclusions, which likely results in the more pronounced differences observed. \\
\indent Notably, in the evaluation presented in Table~\ref{tab:3} using ViT~\cite{ViT}, the difference between our method and existing methods is substantial. Fig.~\ref{fig:A3} compares the concept regions obtained in the experiment. While the existing methods produce noisy regions, our method yields clearer ones. Although WWW~\cite{WWW} sampled images where specific neurons were highly activated during concept matching, these images did not exhibit interpretable semantic consistency for ViT-B/16. Based on these findings, it can be inferred that, particularly for Transformer-based models, representing concepts through the fusion of multiple activation maps is an effective approach.

\begin{table}[tb]
\small
\centering
	\begin{tabular}{|>{\centering}p{5.8em}||>{\centering}p{2.3em}>{\centering}p{2.3em}>{\centering}p{2.3em}>{\centering}p{2.7em}>{\centering\arraybackslash}p{2.3em}|}
		\hline
		& \multicolumn{5}{c|}{AUC (ResNet50 $\times$ Broden)} \\
		\cline{2-6}
		& Object & Part & Color & Material & Avg. \\
		\hline
		CLIP-Dissect & 0.124 & 0.034 & {\bf 0.101} & 0.092 & 0.088 \\
		WWW & 0.114 & 0.034 & 0.091 & 0.064 & 0.076 \\
		CE-FAM & {\bf 0.145} & {\bf 0.044} & 0.100 & {\bf 0.098} & {\bf 0.097} \\
        \hline
        Avg. ${\textrm {AUC}}_{\textrm {high}}$ & 0.220 & 0.089 & 0.197 & 0.183 & 0.172 \\
        Avg. ${\textrm {AUC}}_{\textrm {low}}$ & 0.056 & 0.015 & 0.051 & 0.036 & 0.039 \\
		\hline
        \end{tabular}
\vspace{-8pt}
\caption{AUC results under the same conditions as in Table~1.}
\label{tab:A4}
\end{table}
\begin{figure}[t]
\centering
    \begin{tabular}{p{1\linewidth}}
    \centering
        \begin{minipage}[t]{1\linewidth}
            \centering
            \begin{minipage}[t]{0.3\linewidth}
                \centering
                \subcaption*{ceiling}
                \includegraphics[width=0.7\linewidth]{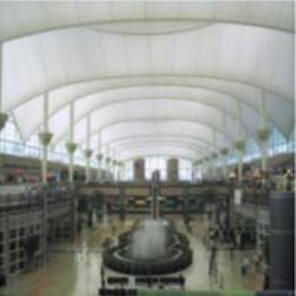}
            \end{minipage}
            \begin{minipage}[t]{0.3\linewidth}
                \centering
                \subcaption*{flowerpot}
                \includegraphics[width=0.7\linewidth]{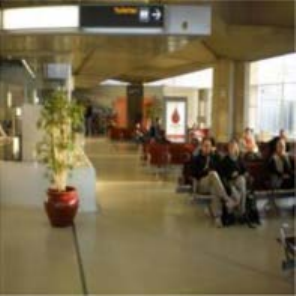}
            \end{minipage}
            \begin{minipage}[t]{0.3\linewidth}
                \centering
                \subcaption*{blue}
                \includegraphics[width=0.7\linewidth]{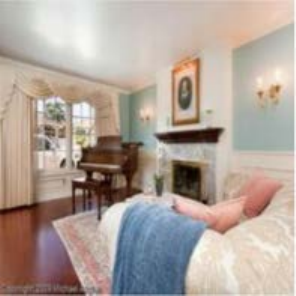}
            \end{minipage} \\
            \vspace{3pt}
            \begin{minipage}[t]{0.3\linewidth}
                \centering
                \includegraphics[width=0.7\linewidth]{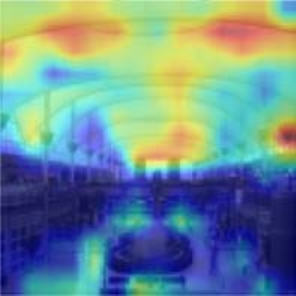}
            \end{minipage}
            \begin{minipage}[t]{0.3\linewidth}
                \centering
                \includegraphics[width=0.7\linewidth]{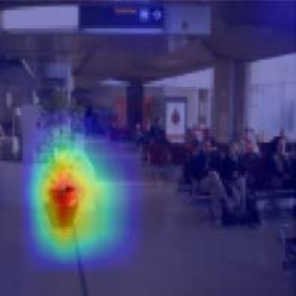}
            \end{minipage}
            \begin{minipage}[t]{0.3\linewidth}
                \centering
                \includegraphics[width=0.7\linewidth]{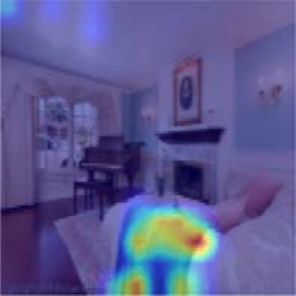}
            \end{minipage}
        \end{minipage}
    \end{tabular}
\vspace{-8pt}
\caption{Concept regions identified by a model trained without Similarity Loss. The images and concept set from Broden~\cite{NetworkDissection} are used.}
\label{fig:A4}
\end{figure}

\subsection{Concept Region for Unseen Concepts}
Our method learns the VLM's embeddings, enabling it to identify concept regions with high accuracy without requiring a predefined concept set. Table~\ref{tab:4} presents results for a model trained without Similarity Loss, and Fig.~\ref{fig:A4} illustrates examples of the concept regions it produces. Despite not being explicitly trained with these concepts as textual inputs, the model successfully identifies the corresponding regions with high accuracy. The better the model mimics the VLM's embeddings, the more effectively it can predict concepts that the VLM has already learned, even without being explicitly provided with those concepts in advance. Note that this setting is close to zero-shot inference but not entirely zero-shot. Although concept labels are not provided during training, image instances are available. Thus, the zero-shot condition applies only to the concept labels.

\section{Advantages and Limitations}
\subsection{Advantages of Our Method}
For a more human-understandable explanation of the reasoning behind image classification, it is important to incorporate the aspects of 5W1H framework (\textit{Who}, \textit{What}, \textit{When}, \textit{Where}, \textit{Why}, \textit{How}) from an ergonomic perspective~\cite{WWW}. The proposed method established a one-to-one correspondence between concept labels, their spatial regions, and their contributions. Our quantitative evaluations demonstrated that the accuracy of these concept regions surpasses that of existing methods. Therefore, the proposed method can indicate \textit{what} concept is involved, \textit{where} it appears, and \textit{how} it contributes to the prediction. It also enables clearer presentation of concept regions by integrating activation maps. While the effectiveness of this idea has been supported by previous studies~\cite{Net2Vec, LoCE, glca}, our work further provides quantitative validation using the newly introduced evaluation metric. In addition, by leveraging VLMs, our method can learn concepts from any image dataset without requiring an annotated dataset. It also demonstrated strong performance in a zero-shot setting with respect to concept labels. With its scalability in handling concepts and flexibility to be applied to various existing models without modification, it provides new value as an explainable AI (XAI) approach that enhances human interpretability.

\subsection{Limitations and Future Directions}
\noindent {\bf Technical limitation.} \indent The performance of our method is limited by the quality of CLIP’s embedding, which has only 512 dimensions. As a result, achieving sufficiently accurate predictions for all concepts is challenging. However, since our evaluation demonstrated superior results compared to existing methods, this limitation is likely to affect most related approaches~\cite{CLIP-dissect, WWW}. Nevertheless, this issue has the potential to be mitigated as VLMs continue to evolve. In fact, when using SigLIP~\cite{SigLIP}, which improves training stability and enhances accuracy by incorporating sigmoid loss and soft labels into CLIP’s training process, improved accuracy of concept regions was observed, as shown in Table~\ref{tab:4}.\\
\indent Moreover, in concept-based XAI, selecting an appropriate concept set is highly important. In this study, Broden~\cite{NetworkDissection} and ImageNet~\cite{ImageNet} were selected as concept sets to ensure a unified evaluation. However, these concept sets may not always be sufficient. If irrelevant concepts are included, they may appear as top-ranked predictions by VLMs, potentially leading to suboptimal explanations. Since efficiently selecting an appropriate concept set is a common challenge in developing human-interpretable XAI, this issue is expected to become a central focus of future research.
\vskip.5\baselineskip
\noindent {\bf Experimental limitation.} \indent In this study, we evaluated the generalizability of our method under various combinations of classifiers, datasets, and VLMs. However, its effectiveness remains unproven outside the evaluated settings. Applying the method to classifiers with different architectures, using datasets from other domains, and optimizing the learning module to support such extensions remain important future directions. Additionally, while concept contributions are measured based on their impact on classification scores, and can be seen as directly representing contribution to some extent, the definition and evaluation of concept contributions can be improved. Clarifying these aspects is essential for achieving more interpretable XAI.

\end{document}